\documentclass[sigconf,nonacm]{acmart}

\AtBeginDocument{%
  }

\copyrightyear{2025}
\acmYear{2025}
\setcopyright{rightsretained}
\setcctype{by}
\acmConference[GECCO '25]{Genetic and Evolutionary Computation Conference}{July 14--18, 2025}{Malaga, Spain}
\acmBooktitle{Genetic and Evolutionary Computation Conference (GECCO '25), July 14--18, 2025, Malaga, Spain}\acmDOI{10.1145/3712256.3726475}
\acmISBN{979-8-4007-1465-8/2025/07}

\begin{document}

\title{When Does Neuroevolution Outcompete Reinforcement Learning \\ in Transfer Learning Tasks?}

\author{
  Eleni Nisioti,
Joachim Winther Pedersen,
  Erwan Plantec,
  Milton L. Montero,
  Sebastian Risi
}
\email{ { enis, jwin, erpl, mile,  sebr}@itu.dk, }
\affiliation{
\institution{IT University of Copenhagen}
\city{Copenhagen}
\country{Denmark}
}

\renewcommand{\shortauthors}{Nisioti et al.}

\begin{abstract}
The ability to continuously and efficiently transfer skills across tasks is a hallmark of biological intelligence and a long-standing goal in artificial systems.
Reinforcement learning (RL), a dominant paradigm for learning in high-dimensional control tasks, is known to suffer from brittleness to task variations and catastrophic forgetting.
Neuroevolution (NE) has recently gained attention for its robustness, scalability, and capacity to escape local optima.
In this paper, we investigate an understudied dimension of NE: its transfer learning capabilities.
To this end, we introduce two benchmarks: a) in \emph{stepping gates}, neural networks are tasked with emulating logic circuits, with designs that emphasize modular repetition and variation b) \emph{ecorobot} extends the Brax physics engine with objects such as walls and obstacles and the ability to easily switch between different robotic morphologies. 
Crucial in both benchmarks is the presence of a \textit{curriculum} that enables evaluating skill transfer across tasks of increasing complexity.
Our empirical analysis shows that NE methods vary in their transfer abilities and frequently outperform RL baselines.
Our findings support the potential of NE as a foundation for building more adaptable agents and highlight future challenges for scaling NE to complex, real-world problems.
  \end{abstract}



\maketitle

\section{Introduction}

% remove biologically plausible and explain that ANNs were used before deep learning: done
Neuroevolution (NE) ~\cite{risi:book25,stanley_designing_2019},
is a rich family of diverse algorithms that can serve as a gradient-free optimization technique for artificial neural networks (ANNs) ~\citep{koza_genetic,stanley_designing_2019}. 
In its early years, genetic programming was considered a sample-efficient and scalable alternative to classical reinforcement learning (RL) algorithms~\citep{stanley_evolving_2002,koza_genetic_nodate, moriarty_efficient_1996}.
Yet, several key innovations in the deep learning era, such as the introduction of novel ANN architectures~\citep{vaswani_attention_2023}, RL objective functions~\citep{schulman_proximal_2017}, and optimizers~\citep{kingma_adam_2017}, have enabled scaling up gradient-based optimization to problems of complexity seemingly unreachable by NE.
Recent studies, however, show that there are cases where NE can outperform RL: evolution strategies can provide compute-efficient approximations of gradients~\citep{salimans_evolution_2017}, genetic algorithms can avoid local optima~\citep{such_deep_2018}, and Quality-Diversity can lead to improved robustness and generalization~\citep{chalumeau_neuroevolution_2023}.
Our work contributes to this accumulating evidence by showing that there is another challenge where certain NE methods can offer an advantage: transfer learning.

Transfer learning, popularized in the field of machine learning through a need for improved data efficiency and generalization, attacks questions as diverse as whether to transfer knowledge, from whom, and in what form~\citep{taylor_transfer_nodate,bengio2009curriculum,khetarpal_towards_2022}.
Here, we are interested in a rather concrete flair of transfer learning: \textit{can an algorithm master a difficult task by moving through sub-tasks of increasing difficulty?} 
This question can be particularly interesting for NE, which:
a) can enable architecture search and, therefore, promote the emergence of structural features such as modularity, repetition, and hierarchy~\citep{gruau_automatic_1994, gruau_genetic_1992,risi_evolving_2010, stanley_hypercube-based_2009, stanley_evolving_2002, stanley_taxonomy_2003}, that have been linked to alleviating catastrophic forgetting~\citep{ellefsen_neural_2015};    
b) may maintain a population of solutions and employ explicit mechanisms for diversity-preservation~\citep{chalumeau_neuroevolution_2023, such_deep_2018,stanley_evolving_2002}, a capacity that has been linked to generalization but, we posit, could potentially also alleviate forgetting well-performing solutions.

%b) optimize a different objective to RL and (note that this is true even for NE methods that approximate gradients~\citep{salimans_evolution_2017}), where catastrophic forgetting has been linked to the non-stationarity introduced by task-switching and the inherent plasticity-stability dilemma~\citep{khetarpal_towards_2022}

% explain better than ecorobot is general but here we focus on transfer learning: done
To investigate this, we introduce two benchmarks. The first one, \emph{stepping gates}, contains tasks that require emulating the behavior of digital circuits consisting of different logic gates.
Here, we instantiate two circuits: the classical \emph{N-parity}, and our 
newly introduced \emph{Simple ALU}, where the circuit supports switching within an instruction set of 16 simple, and composite logic functions, implementing a simplified version of an Arithmetic Logical Unit (ALU). 
N-parity aims at testing the ability to repeat identical modules, while the Simple ALU further requires repetition with modification (see Section \ref{sec:benchmark_stepping_gates} for a closer analysis of this benchmark).
The second one, \emph{ecorobot}, extends the physics-based engine Brax \citep{freeman_brax_2021} with the presence of objects such as walls and obstacles and the ability to easily switch between different robotic morphologies.
In this way, we can disentangle the challenge of locomotion from a diversity of other behavioral challenges, such as maze navigation and obstructed locomotion, (as our results show this can shed light into limitations of different approaches). 
While both benchmarks can capture diverse challenges, here we focus on tasks that require transfer learning, which we impose by decomposing tasks into levels presented to the agent in order of increasing difficulty (we make a small exception to study avoidance of local optima in Section \ref{sec:results_indirect}.)

% here I need to motivate map-elites better (as in explain that it is not an alternative, it could include both direct and indirect encodings. maybe in overall introduce the algorithms better. and then mention in discussion that future work could consider combinations of these methods and explain which we deem to be the most promising
Our empirical study aims to reveal how choices in the design of NE approaches influence their performance in the face of certain challenges. 
For this reason, we benchmark methods that differ across three orthogonal dimensions: 
a) whether they evolve the structure of the ANN or just the weights.
For example, we show that NEAT~\citep{stanley_evolving_2002}, an algorithm that evolves both structure and weights, performs well across many transfer learning tasks (see sections \ref{sec:results_stepping_gates} and \ref{sec:results_ecorobot}); 
b) whether they are direct or indirect, i.e., whether evolution searches directly in the space of ANNs or a genotype to phenotype (GP) map is employed.
Here, we show that HyperNEAT~\citep{stanley_hypercube-based_2009}, an indirect encoding, is disadvantaged at transfer learning but superior at avoiding local optima (see Section \ref{sec:results_indirect}).
c) whether they explicitly search for diversity or not. Here, we show that MAP-elites~\citep{mouret2015illuminating,chalumeau_neuroevolution_2023}, an algorithm excelling in deceptive tasks, is, perhaps surprisingly, at a disadvantage when we introduce a slight modification to the task that requires transfer learning (see Section \ref{sec:results_stepping_stones}).
We also benchmark PPO, a state-of-the-art RL algorithm ~\citep{schulman_proximal_2017}, and goal-conditioned PPO ~\citep{colas_autotelic_2022}, a simple variant designed to 
alleviate non-stationarity by conditioning the policy on the current task.
Both methods exhibit sub-par transfer learning abilities.
Finally, we show that many of the aforementioned benefits of NE disappear when the task remains unchanged, but the robotic morphology becomes more complex.

Overall, our study confirms a variety of known results from the NE and RL literature, introduces new insights into differences between NE approaches, and offers new benchmarks aimed at supporting future work towards scaling up NE to richer behavioral challenges. 
We open-source our benchmarks in two online repositories~\footnote{\href{https://github.com/eleninisioti/ecorobot}{https://github.com/eleninisioti/ecorobot},\href{https://github.com/eleninisioti/stepping_gates}{https://github.com/eleninisioti/stepping\_gates}}.
We also accompany our empirical analysis with a repository for reproducing our experiments ~\footnote{\href{https://github.com/eleninisioti/neuroevolution_in_transfer_learning}{https://github.com/eleninisioti/neuroevolution\_in\_transfer\_learning}} and a website offering visualizations of behavioral trajectories~\footnote{\href{https://sites.google.com/view/neuroevo-transfer/home}{https://sites.google.com/view/neuroevo-transfer/home}}.

%when faced with the challenge of transferring skills within a task of increasing complexity, NEAT outperforms both RL and other NE algorithms, such as evolution strategies \citep{salimans_evolution_2017} and MAP-Elites \citep{chalumeau_neuroevolution_2023}.
%In particular, we experiment with PPO, SAC and a goal-conditioned version of PPO and observe that they can only reach a small number of initial stepping stones.
%This agrees with previous studies that have investigated the problem of catastrophic forgetting in RL \citep{ecoffet_first_2021,schopf_hypernetwork-ppo_2022}.
%On the other hand, we show that MAP-elites, an approach particularly successful in hard exploration problems, is outcompeted by NEAT: as this method reaches the final solution without the need to go through the stepping stones, it accrues less rewards during the episode.
%Our studies also aims to show the limitations of current NE approaches: when increasing the difficulty of the tasks by introducing more complex robotic morphologies, NEAT no longer succeeds.

%Our study shows that NE algorithms that evolve both the architecture and weights of ANNs are a particularly promising research direction towards the objective of artificial systems with transferrable skills  and that more effort is required to scale them up to problem settings of real-world complexity.  

% maybe separate

\section{Background and related works}\label{sec:background}
The design space of NE can be divided into: a) the evolutionary process, which includes mutation, selection, and cross-over operators that search through the genomic space and can be extended with additional diversity-preservation mechanisms such as Quality-Diversity optimization and speciation~\citep{stanley_evolving_2002, mouret2015illuminating, chalumeau_neuroevolution_2023}, and b) the process that maps genomes to phenotypes. From the perspective of this genotype-to-phenotype (GP) map, NE approaches can be divided into \emph{direct} and \emph{indirect} ones.
In the former category, evolutionary search is performed directly in the space of ANNs and may optimize both the architecture and weights~\citep{stanley_evolving_2002, gruau_automatic_1994, risi_evolving_2010} or just the weights of the networks~\citep{stanley_hypercube-based_2009, salimans_evolution_2017,such_deep_2018}.
Discovering optimal architectures is a high-dimensional, non-differentiable problem where NE could potentially show its true advantage over gradient-based methods~\cite{papavasileiou_systematic_2021,stanley_evolving_2002}.
Yet the most well-performing algorithms today avoid optimizing the architecture~\citep{salimans_evolution_2017,such_deep_2018,chalumeau_neuroevolution_2023} as, doing so significantly increases the search space size and introduces the additional challenge of designing mutation and cross-over operators that do not render phenotypes non-functional.
%the presence of varying architectures creates the challenge of defining mutation and cross-over operators that do not render phenotypes non-functional.
To address these challenges, NEAT~\cite{stanley_evolving_2002}, an algorithm where networks start out small and grow gradually with evolutionary time, introduced a number of algorithmic innovations and spawned a long lineage of approaches~\cite{papavasileiou_systematic_2021}.

By interleaving a non-linear mapping from a, traditionally lower-dimensional, genomic space to the phenotypic space, indirect encodings have been associated with benefits such as compression and regularization ~\cite{shuvaev_encoding_2024}, ability to induce structural features such as modularity and repetition~\cite{stanley_hypercube-based_2009,ellefsen_neural_2015}, robustness to phenotypic noise~\cite{najarro_hypernca_2022} and improved navigability of the genomic space~\cite{greenbury_structure_2022}.
Notwithstanding these impressive advantages and highly contingent on their particular implementation, indirect encodings can be hard to optimize due to inducing chaotic or degenerate optimization landscapes~\cite{thomson_understanding_2024, nisioti_growing_2024}.
Our study contributes to this empirical knowledge the observation that indirect encodings can more easily avoid local optima but, on the downside, fail to transfer skills across tasks.

\begin{figure*}
    \centering
    \includegraphics[width=0.75\linewidth]{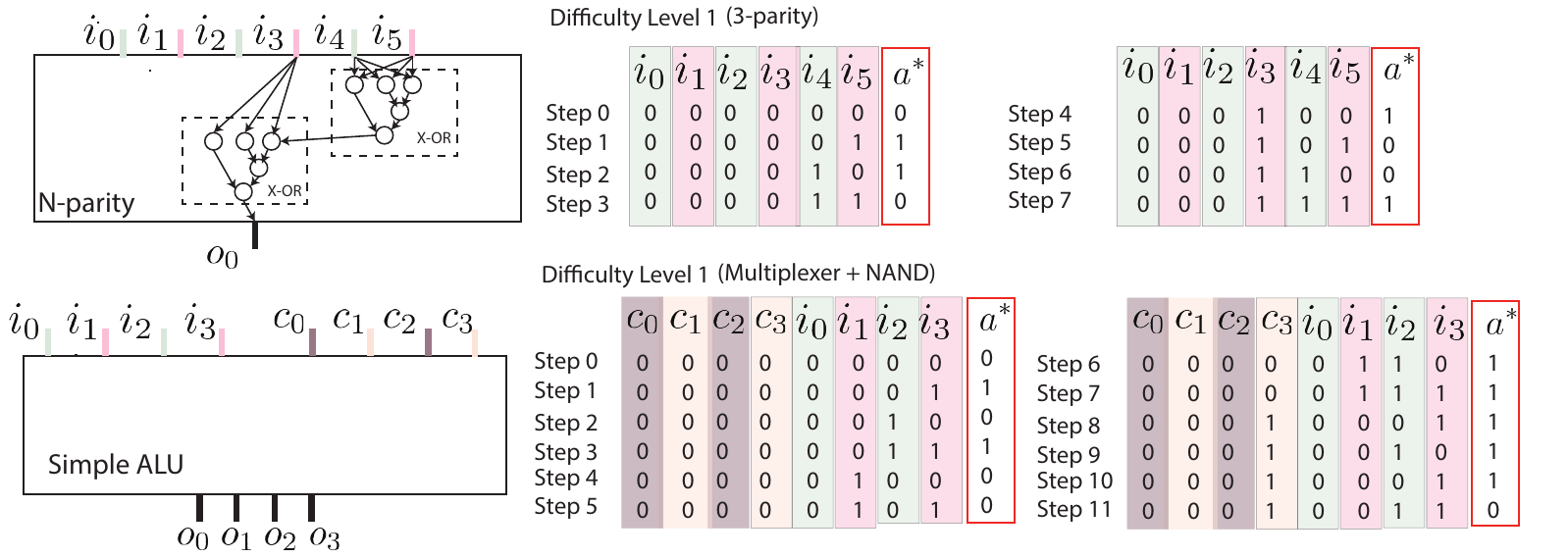}
    \caption{Illustration of the stepping gates benchmark. \normalfont (top) N-parity is a task with N input bits and 1 output bit where the first level starts with two input bits and each next level adds an extra bit. In this example, with a task of 3 bits at level 1, a full episode will go through all 8 combinations of the 3 active bits. Inactive bits are set to 0 and the optimal action for each step is depicted inside the red rectangle. On the left we illustrate a solution to the 3-parity problem that leverages a solution to 2-parity. (bottom) The Simple ALU task has 4 inputs, 4 output, and 4 control bits. The episode length and number of active inputs depend on the current level. At level 1, the task requires implementing a multiplexer and a NAND gate. A single control bit is used to switch between the two.}
    \label{fig:stepping_gates_tasks}
\end{figure*}

Recently a number of studies have placed NE under close scrutiny in terms of its ability to out-compete its main competitor in control tasks: RL~\citep{chalumeau_neuroevolution_2023,salimans_evolution_2017, such_deep_2018}.
For example, genetic algorithms can better avoid local optima~\citep{such_deep_2018}, evolution strategies can explore more efficiently~\citep{salimans_evolution_2017}, and Quality-Diversity optimization can lead to improved robustness and generalization~\citep{salimans_evolution_2017}.
Our study suggests that transfer learning is another such frontier.
The inability of RL to deal with this challenge is a long-established observation that has been attributed to the non-stationarity introduced when switching between tasks~\citep{khetarpal_towards_2022,huang_continual_2021,ecoffet_first_2021}.
To the best of our knowledge the study of transfer learning for NE has been limited to a small diversity of tasks and algorithms~\citep{ellefsen_neural_2015}.
Here, we aim at scaling it up across a wider spectrum.

With more powerful NE algorithms, tasks have become increasingly complex, evaluating agents in high-dimensional, stochastic spaces.
For example, tests developed for evaluating animal cognition can be used to drive artificial systems toward animal-level intelligence~\citep{crosby_animal-ai_2019, lucas_articulated_2024}.
These tasks often focus on specific cognitive abilities like planning and memory, using simplified agent morphologies~\citep{voudouris_direct_2022}.
On the other hand, benchmarks involving complex robotic morphologies often focus on locomotion~\citep{freeman_brax_2021}, though recent works also evaluate tasks combining locomotion with object interaction in 3D~\citep{lim_accelerated_2022} and 2D~\citep{matthews_kinetix_2024}.
Ecorobot, a benchmark we open-source in this study, facilitates these setups by offering a variety of morphologies and tasks.
An additional consideration when it comes to benchmarks is computational complexity, with modern environments relying on GPU-accelerated game engines~\citep{crosby_animal-ai_2019, lucas_articulated_2024} or JAX-based physics frameworks~\citep{grillotti_kheperax_2023, freeman_brax_2021, lim_accelerated_2022}. Stepping gates and ecorobot are written in JAX for efficient, GPU-based end-to-end training.

\section{Benchmarks}

\subsection{Stepping gates}\label{sec:benchmark_stepping_gates}
Digital circuits are a long-standing domain for NE studies due to their simplicity (searching in the space of small, discrete-weight networks was manageable for genetic programming and grammar-based approaches~\citep{koza_genetic_1991, gruau_genetic_1992}) and structural properties (circuits are by design modular, regular and hierarchical, enabling the study of architecture search).
Inspired by this tradition, this benchmark defines tasks that add to the above benefits the presence of a curriculum: a task consists of a series of levels corresponding to different circuits, with circuits of later levels being compositions of earlier ones. 
Each circuit corresponds to a simple logical operation (such as NAND and logical shift left).
In this study we consider the following two tasks.

\subsubsection{N-parity} The agent needs to compute the parity of $N$ bits. There are $N-1$ levels of difficulty, where the first level requires computing the parity of two bits (X-OR), each level adding an extra bit.
The agent can move to the next level only when it has solved the current one.
This task captures the property of modularity and repetition: as we illustrate in the top of Figure \ref{fig:stepping_gates_tasks}, a solution found for the first level (X-OR) can be copied and connected hierarchically for subsequent levels.

\subsubsection{Simple ALU} The network needs to simulate a simplified version of an Arithmetic Logical Unit that can switch between 16 logical operations using 4 control bits.
There are 9 levels of increasing difficulty: multiplexer, NAND, NOT, AND, XOR, increment by one, decrement by one, logical shift left, and logical shift right.
While this ordering does not necessarily lead to a monotonic increase in difficulty and is just one of many possible options, it does capture the intuitive properties that some logical operations form the basis for others.
As with N-parity, the agent can move to the next level only when it has solved the current one, with the added complexity that to solve a level the agent needs to emulate all previous ones (so that the final level implements the ALU in its entirety).
This task captures modularity and repetition with variation (for example the addition of a single edge can convert a NAND gate into a NOT).

This benchmark can be employed in supervised learning, RL, where we formulate them as environments with an episode of a single step and randomly sample from the observation space at each episode, and NE, where we consider an episode of length $N^2$ that consists of all possible inputs.
Here we benchmark the RL and NE set-ups.
We provide an illustration of how an episode unfolds in these two tasks in Figure \ref{fig:stepping_gates_tasks}.

\begin{figure*}
    \centering
    \includegraphics[width=0.75\linewidth]{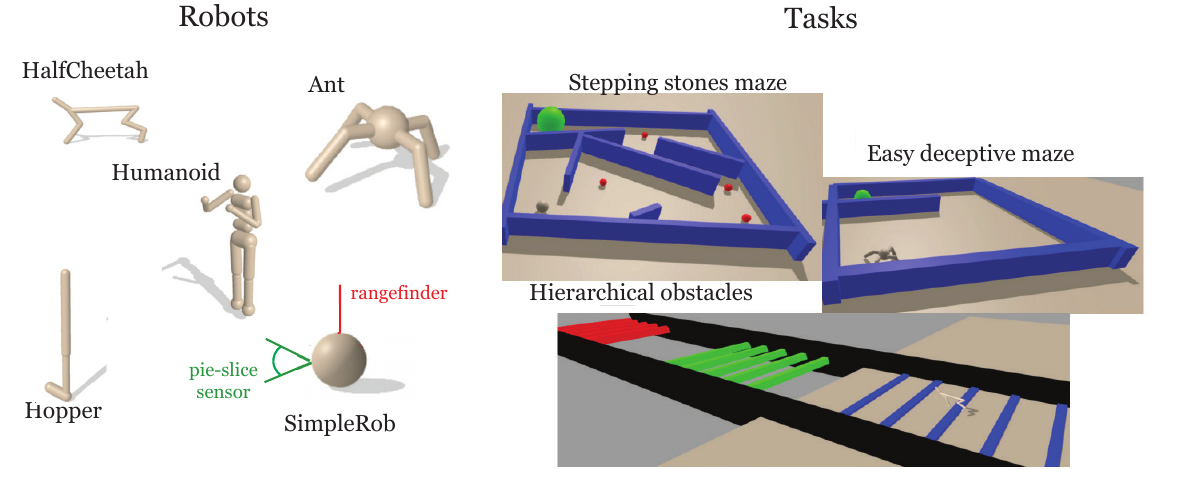}
    \caption{Illustration of the ecorobot benchmark. \normalfont An environment consists of a robot and a task. (left) We currently support different Brax robots and a simpler robot, SimpleRob. Robots can be equipped with rangefinders and pie-slice sensors. (right) A task consists in a choice of objects and a reward function. We have implemented tasks that test for different behavioral challenges, with the Stepping stones maze and Hierarchical obstacles tasks being specifically designed to test for transfer learning.}
    \label{fig:ecorobot_tasks}
\end{figure*}

\subsection{Ecorobot}\label{sec:benchmark_ecorobot}

Ecorobot is a benchmark that enables the evaluation of agents in environments requiring locomotion and navigation.
Built on top of the Brax physics engine~\citep{freeman_brax_2021}, it offers an interface for designing tasks where robots interact with items such as walls, obstacles, and food.
We have provided a set of tasks largely inspired by previous NE studies that test for specific behavioral skills.
An environment in ecorobot is a combination of a robot and a task.
% say that these appear in the figure

\subsubsection{Robots}

Robots are three-dimensional compositions of simple geometrical shapes with joints controlled by continuous actuators.
In addition to the locomoting robotic morphologies offered by Brax (Ant, HalfCheetah, Hopper, Humanoid), we have implemented a simple robot that is a plane sphere controlled by four actuators enabling movement on the x-y plane (SimpleRob).
A robot can be equipped with two types of sensors that can be positioned along its periphery: rangefinders indicate the distance to the closest item along the direction perpendicular to their attachment point on the robot while pie-slice sensors indicate the presence of an item within a certain angle around them ~\footnote{Sensors were introduced very recently in the mujoco-xla backend so this functionality may be missing from other similar simulators}.
We illustrate the different robots and sensor placement on the left of Figure \ref{fig:ecorobot_tasks}.

\subsubsection{Tasks}
A task is a Brax environment extended with the presence of items such as walls, obstacles, and food, as well as its custom reward function.
We have implemented different tasks that test for locomotion and navigation under different challenges.

\paragraph{Locomotion} This task requires maximizing velocity along the robot's forward (positive x) direction and corresponds to the original Brax environment.

\paragraph{Deceptive maze} The agent needs to navigate through a maze to reach a target food located at the top left corner of the maze starting from the bottom left.
The reward is the distance traveled towards the goal so that an optimal trajectory incurs a reward of 1.
The agent is equipped with five rangefinder sensors that detect walls.
This task contains a local optimum: maximizing the current reward leads to getting trapped at the northern wall.
We have implemented two versions of the deceptive maze: easy and difficult (Figure \ref{fig:ecorobot_tasks} presents the easy deceptive maze while the difficult one looks identical to the Stepping stones maze with the difference that the red stepping stones are absent). 
This task is commonly used in studies of local optima~\citep{chalumeau_neuroevolution_2023,lehman_abandoning_2011}.

\paragraph{Maze with stepping stones} This task is identical to the deceptive maze, except for the reward function: the maze includes a number of locations, referred to as stepping stones, that create a trajectory leading to the food (we visualize them as spherical objects but they are undetectable by the agent).
The reward the agent receives at each step is the number of stones it has crossed (there is a minimum distance of 0.1 for detecting crossing) plus the distance to the next stone, normalized in the 0-1 range. 
Crossing the stones needs to happen in the correct order to incur the desired reward.
Upon reaching the food, the agent receives a high reward (1,000).
The optimal behavior in this task is crossing all stones and reaching the target, which incurs a fitness of about 10,000.
This task has been used in past NE studies interested in how methods can build on stepping stones~\citep{stanley_hypercube-based_2009,risi_evolving_2010}.

\paragraph{Hierarchical obstacles} The agent needs to maximize its velocity in its forward direction while stepping over a series of obstacles.
The difficulty of crossing an obstacle is controlled by setting its height.
Obstacles are arranged in three levels of increasing difficulty. 
There is also an initial, obstacle-free area.
A simpler version of this task that included a single level of difficulty was employed in previous works~\citep{chalumeau_neuroevolution_2023,eysenbach_diversity_2018}.
%Our extension aims to make this need for hierarchical learning clearer.

Thus, the tasks in ecorobot that test for transfer learning abilities are the maze with stepping stones and the hierarchical obstacles.
In the former, the agent needs to remember the series of stepping stones it has discovered in the past.
In the latter, the agent can first learn to locomote without obstacles and, then, adjust to obstacles of increasing difficulty.
All tasks have an episode length of 1,000 steps.

\section{Methods}\label{sec:methods}
We evaluate a variety of NE and RL algorithms with the aim of capturing different dimensions of their design space.
For each algorithm, we provide a high-level description (that relies on concepts we described in Section \ref{sec:background}) and the hyperparameters we employed in Appendix \ref{app:hyperparams}).
For more algorithmic details, we refer readers to the works that originally introduced these algorithms and provide an online repository with our implementations~\footnote{\href{https://github.com/eleninisioti/neuroevolution_in_transfer_learning}{https://github.com/eleninisioti/neuroevolution\_in\_transfer\_learning}}.

\emph{NEAT}~\citep{stanley_evolving_2002} is a direct encoding that optimizes both the architecture and weights of an ANN.
It employs a population of ANNs that undergoes: a) mutations, which can add or remove nodes and edges and add random noise on weights b) cross-over, where differing architectures are appropriately aligned through the use of historical markings and, c) selection, where the fittest individuals reproduce and an elite is chosen that does not undergo mutations.
NEAT networks start out with a minimal size and gradually grow across generations. 
We employed the implementation of NEAT provided in the tensorneat library~\citep{wang_tensorized_2024} with the hyperparameters detailed in the appendix.

\emph{HyperNEAT}~\citep{stanley_hypercube-based_2009} is an indirect encoding that evolves a network using NEAT that predicts the weights of another network, used as a policy network.
NEAT optimizes the structure, weights and activation functions of the first network, called a Compositional Pattern-Producing Network (CPPN), which accepts as inputs the spatial location of two neurons in the policy network and outputs the weight value of the edge connected them.
Thus, this GP map is a forward pass through the CPPN.
The use of certain activation functions can induce certain regularities in phenotypes, such as symmetry and repetition.  
A potential challenge when employing HyperNEAT is that the spatial locations of the neurons of the policy network need to be determined manually .
In tasks where geometry plays a role, this positioning can take advantage of our geometry of the task (see Figure \ref{fig:hyperneat_substrate} in Appendix \ref{app:hyperparams} for an illustration of how we placed them for SimpleRob).

%for example, in a robot locomotion task right/left and front/back symmetry can be taken into account when placing sensory input neurons and motor output neurons.
%Thus, in ecorobot tasks, we have placed neurons appropriately
%For stepping gates we have simply placed them in a feedforward architecture with a single hidden layer.
%We provide hyperparameters for this method in Table \ref{tab:hyperneat_hyper} of the appendix. 

\emph{Covariance Matrix Adaptation Evolution Strategy} (CMA-ES)\citep{hansen_cma_2023} is a direct encoding that optimizes the weights and biases of a neural network.
It belongs to the broader family of evolution strategies, which operate by sampling candidate solutions from a multivariate normal distribution and iteratively updating this distribution based on performance. CMA-ES adapts the covariance matrix of the distribution over generations, allowing it to capture dependencies between parameters and perform efficient search in high-dimensional, non-convex landscapes. 
We employed the implementation provided in the evosax library~\citep{evosax2022github}.

 \emph{MAP-Elites}~\citep{mouret2015illuminating} is an NE algorithm belonging to the family of Quality-Diversity optimization~\citep{pugh_quality_2016}, a paradigm emphasizing that it is both performance and diversity that matter in evolution. 
 In Map-Elites, solutions are attributed a fitness and a set of behavioral descriptors lying in a pre-defined task-specific behavior space.
 This behavior space is partitioned into discrete cells which are populated with the best performing individuals whose behavioral descriptors lie in the cell domain.
 At each generation, a random cell is chosen, its associated genome is mutated and the newly created phenotype is evaluated in terms of its performance and behavioral characterization.
 Then, the solution is added to the associated cell provided it has higher fitness than the solution inhabiting it or the cell is empty.
 A MAP-Elites algorithm can employ different evolutionary operators and GP maps, it should therefore be viewed as a set of algorithms that introduces the orthogonal dimension of diversity optimization.
 In our study, we have employed a simple version of MAP-Elites that uses a genetic algorithm and a direct encoding, as our aim is not to discover the most well-performing variant but to study how the consideration of diversity affects transfer learning.
 We employed the implementation provided in the QDax library~\citep{chalumeau2024qdax}.
 
\emph{Proximal Policy Optimization} (PPO)~\citep{schulman2017proximalpolicyoptimizationalgorithms} is an RL algorithm belonging to the family of policy-gradient methods.
 It balances performance and stability by constraining policy updates through a clipped surrogate objective that prevents large changes.
 %It employs two neural networks: an actor, which outputs actions given states, and a critic, which estimates the value function that guides learning.
 PPO benefits from sample efficiency, making it a popular choice for continuous control and high-dimensional tasks and, in contrast to most NE approaches, is very sensitive to the choice of hyperparameters~\citep{such_deep_2018}.
 We employed the implementation provided in Brax \citep{freeman_brax_2021}. 

\emph{Goal-conditioned PPO}
A simple variant of PPO is one where the policy network accepts as input a vector indicating the current goal, in addition to the current environment observation. 
~\citep{colas_autotelic_2022}.
Such an approach can theoretically alleviate the problem of non-stationarity present in transfer learning tasks, as changing goals are the causal factors of non-stationarity ~\citep{kaelbling_learning_1993, khetarpal_towards_2022}.
We employed the same implementation with PPO.
A challenge with this approach is the need for defining meaningful goals for the different levels of a task.

As a brief note regarding how we performed hyperparameter tuning: NEAT and HyperNEAT were tuned in the 6-parity task, PPO and goal-conditioned PPO  were tuned independently for the stepping-gates (6-parity) and ecorobot tasks (locomotion with the ant), CMA-ES employed its default hyperparameters and the hyperparameters for MAP-Elites were taken from~\citep{lim_accelerated_2022}.

\section{Results}
The goal of the present study is to highlight how the tasks we have introduced pose different and meaningful challenges to the methods discussed in Section \ref{sec:methods}. 
We do not aim at an exhaustive evaluation of all combinations of tasks and methods (some of them would not be trivial, as goal-conditioned RL requires the notion of goals and MAP-Elites behavioral descriptors, both not readily available for all tasks).
We perform 10 independent trials for each condition and highlight in barplots the p-value for statistically significant comparisons between pairs of methods
\footnote{Since our data did not follow a normal distribution, we used the non-parametric Kruskal-Wallis test to assess overall group differences, followed by Mann-Whitney U tests for post-hoc pairwise comparisons. The number of asterisks indicates the range of p-values: 0.005 < * < 0.05, 0.0005 < ** < 0.005, 0.00005 < ** < 0.0005.}.

\subsection{Analysis in stepping gates benchmark}\label{sec:results_stepping_gates}
In this benchmark we quantify performance as the percentage of all possible combinations of inputs for which a method provided the correct output in a given episode. 
We refer to this value as \emph{success}.
As described in Section \ref{sec:benchmark_stepping_gates},
a task starts at the first level and moves to the next level only once the agent has achieved perfect success in the current one.
Thus, the number of possible inputs depends on both the task and the current level.
For example, when in the third level of the N-parity task, a success of 1 indicates that the method has successfully predicted all $2^3$ combinations of 3 inputs bits.

\begin{figure}
    \includegraphics[width=0.9\columnwidth]{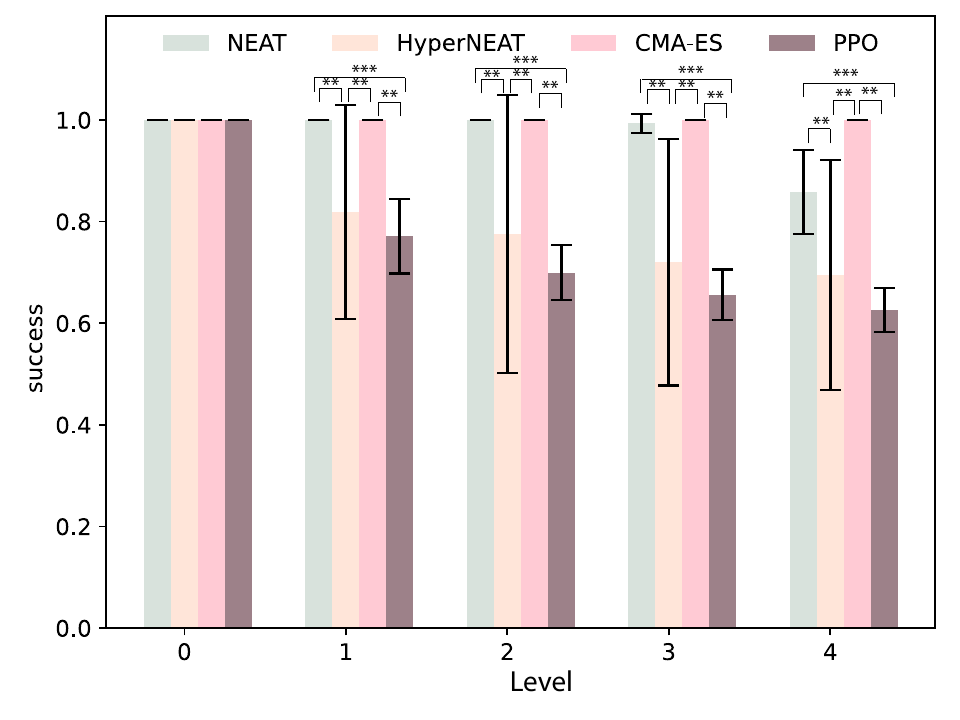}
        \includegraphics[width=0.9\columnwidth]{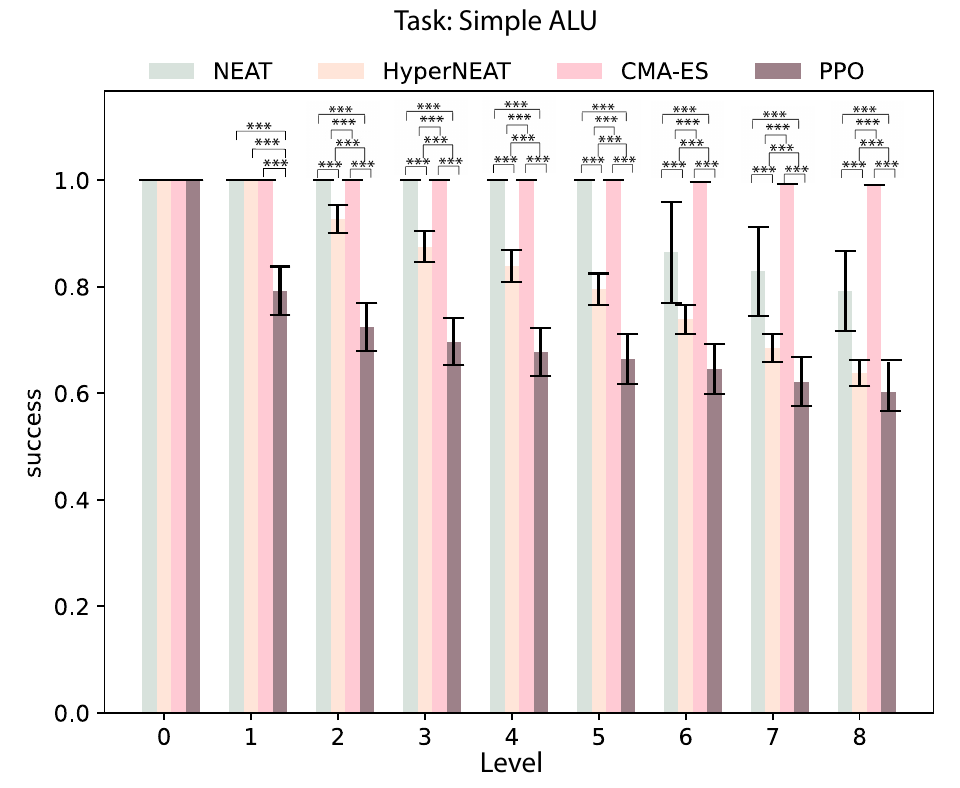}
    \caption{Success in the stepping-gates benchmark with levels of increasing difficulty: \normalfont (top) N-parity (bottom) Simple ALU  }
    \label{fig:results_stepping_gates}
\end{figure}

\begin{figure}
    \centering
    \includegraphics[width=0.9\linewidth]{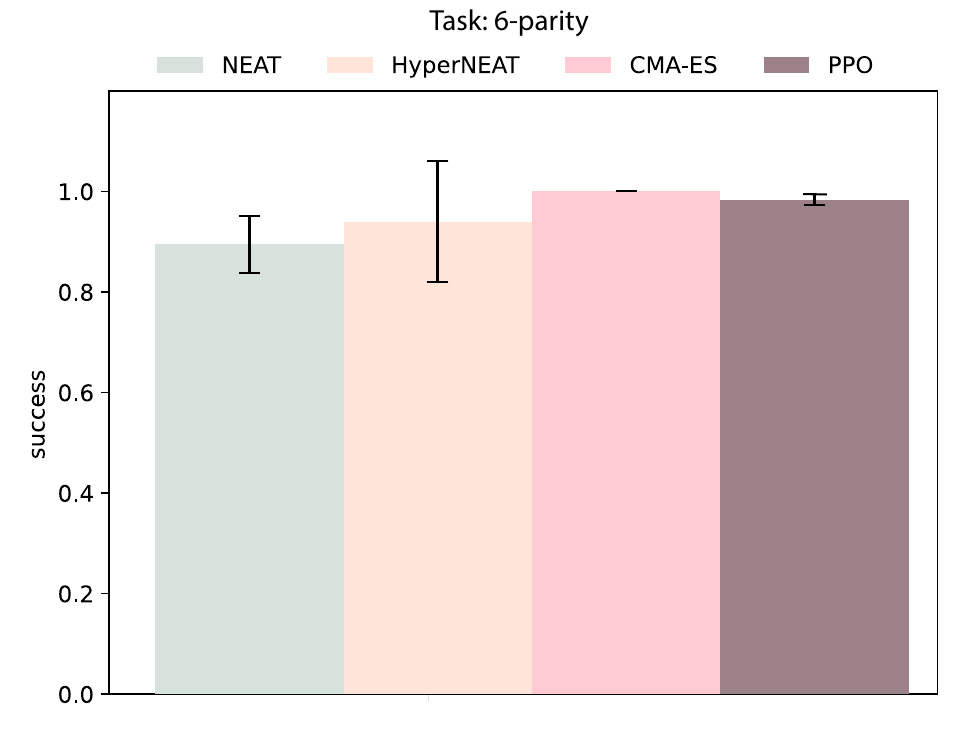}
    \caption{Ablation for the N-parity task. \normalfont  When directly solving 6-parity (without going through the intermediate levels) PPO solves the task while NEAT's performance degrades slightly.}
    \label{fig:parity_ablation}
\end{figure}

Figure \ref{fig:results_stepping_gates} presents the performance of different methods in the N-parity (top) and Simple ALU (bottom) tasks.
%As this task does not readily lend itself to defining goals and behavioral descriptors, we did not benchmark goal-conditioned PPO and MAP-elites (but we do so in subsequent tasks).
The two tasks lead to similar conclusions:
a) direct encodings perform best in this benchmark.
CMA-ES successfully solves all levels for both tasks, while NEAT solves N-parity but stops at level 5 of the Simple ALU task.
The latter is not surprising, as the Simple ALU task is more challenging: it has a larger observation/action space and requires emulating more complex logic operations  
b) HyperNEAT initially makes some progress, but its performance quickly drops with increasing levels.
This agrees with previous works showing that, due to its indirect nature, this encoding has trouble building on top of previous solutions \citep{risi_evolving_2010}
c) PPO  only reached the first level of both tasks. To understand this behavior we performed an ablation for the N-parity task: instead of progressively going through all levels we directly attempt to solve the 6-parity problem.
In Figure \ref{fig:parity_ablation}, we observe that, now, PPO achieves perfect success. 
This suggests that the issue does not lie in the capacity of PPO but in its inability to progress through the levels of our curriculum.
As we discussed in Section \ref{sec:background} this is a known limitation of RL methods.
Interestingly, it is NEAT that now struggles with the task's lack of curriculum.

\subsection{Analysis in ecorobot benchmark}\label{sec:results_ecorobot}
For this benchmark, we analyze the rewards and behavioral trajectories in tasks that challenge methods in terms of transfer learning, locomotion complexity, and avoidance of local optima.

\subsubsection{Transfer learning}
\label{sec:results_stepping_stones}
As we saw in Section \ref{sec:benchmark_ecorobot}, we have introduced two tasks in ecorobot aimed at the study of transfer learning: the maze with stepping stones and locomotion with obstacles.
To disentangle the challenge of locomotion from the challenge of transfer learning, which we deem to be orthogonal, we first study the SimpleRob robot in the aforementioned maze (we will move to complex morphologies right after).
In addition to the methods we studied in the stepping gates benchmark, we also study:
a) goal-conditioned PPO, where goals are the two-dimensional coordinates of the currently active stepping stone
b) MAP-Elites, where, similarly to a previous study in deceptive mazes, we employ two-dimensional coordinates as behavioral descriptors \citep{lim_accelerated_2022}.

\begin{figure}
    \centering
    \includegraphics[width=0.9\linewidth]{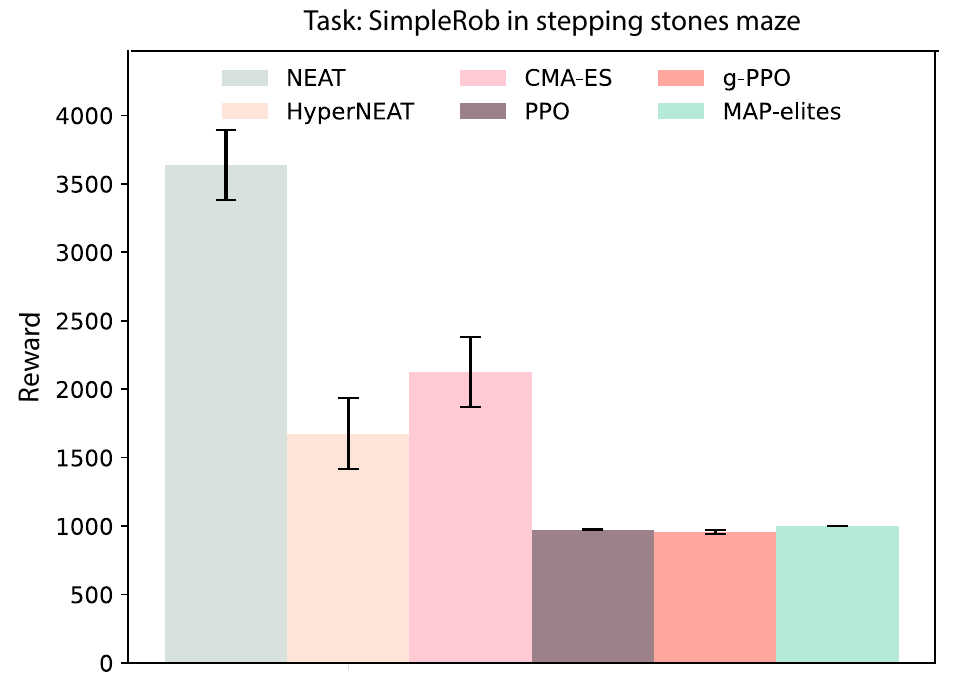}
    \caption{Comparing the ability of methods to progress through stepping stones in the maze {\normalfont NEAT and CMA-ES perform best, PPO reaches only the first stepping stone while MAP-elites collects low rewards due to reaching the end of the maze without collecting the stones}} 
    \label{fig:stepping_rob}
\end{figure}

\begin{figure}
    \centering
    \includegraphics[width=0.9\linewidth]{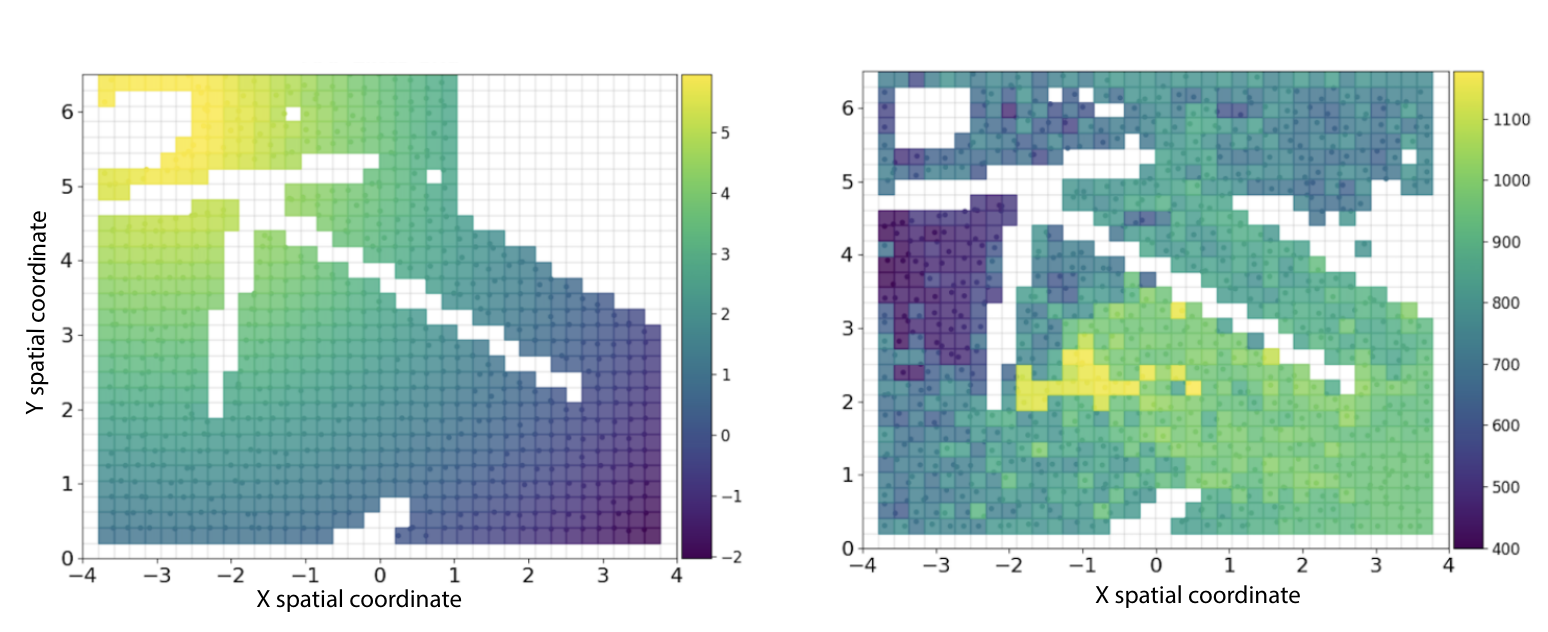}
    \caption{\normalfont \textbf{Left:} Map of elites in the deceptive maze. \textbf{Right:} Map of elites in the maze with stepping stones. 
    While coverage is optimal in both cases, rewards in the stepping stones maze are low as an optimal trajectory requires traversing through the stepping stones in the right order.
    %The behavior descriptors are the x- and y-coordinates of the agent at the last time step of the episode. While the MAP-elites algorithm produces agents that visit most of the maze in both cases, the agents do not go to all the stepping stones in the correct order in the maze with stepping stones. The most rewarded individuals are thus the ones that end the episode around the first stepping stone in the maze.
    }
    \label{fig:mapelite_maze}
\end{figure}

In Figure \ref{fig:stepping_rob}, we present the rewards accrued by different methods (we also provide visualizations of behaviors that are a necessary basis for our discussion \footnote{\href{https://sites.google.com/view/neuroevo-transfer/home}{https://sites.google.com/view/neuroevo-transfer/home}}):
a) NEAT accrues the highest rewards, reaching on average five out of seven stepping stones
b) HyperNEAT and CMA-ES manage to reach about three stepping stones
c) both RL methods get stuck at the first stepping stone. 
d) MAP-Elites performs equally bad but exhibits a drastically different behavior. As we see on the right of Figure \ref{fig:mapelite_maze}, the agent has explored the whole maze (in contrast to PPO that only reached the first stepping stone). To put this behavior into perspective we juxtapose it to the classical deceptive task (that only differs from our maze in the absence of stepping stones).
In this task, the agent's extensive exploration leads to finding the optimal policy: reaching the target on the top left.
In our task, however, reaching the final target is not sufficient: unless the agent visits stepping stones in the right order, reaching the target will incur minimal rewards.

\begin{figure}
    \centering
        \includegraphics[width=0.9\linewidth]{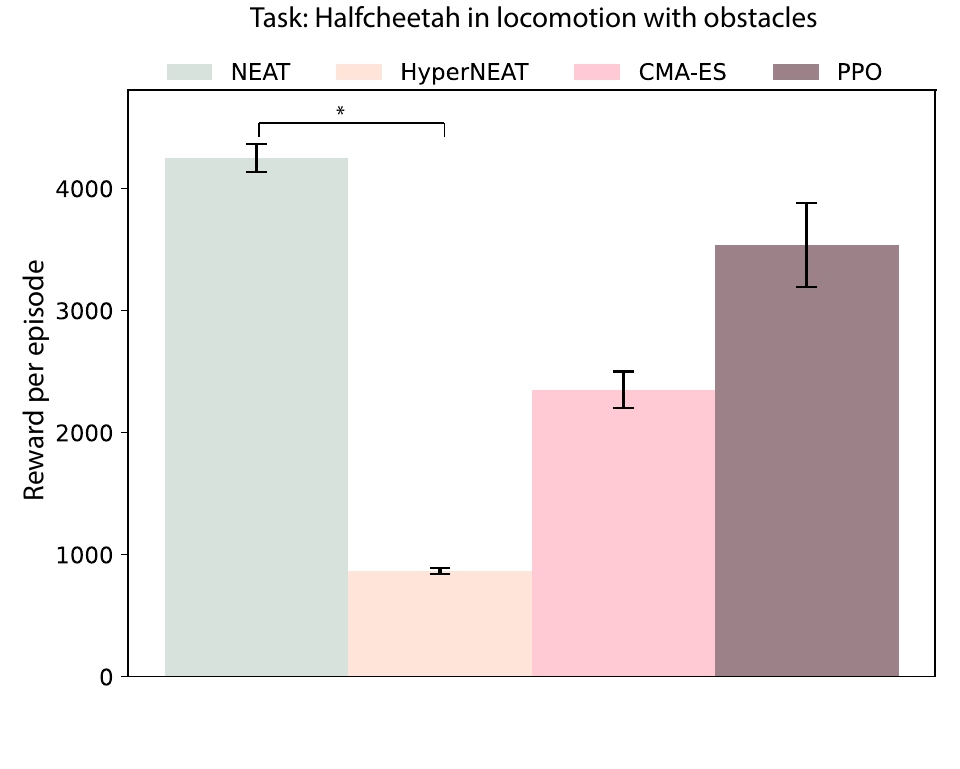}
    \caption{NEAT performs best at locomoting over obstacles of increasing height. {\normalfont By inspecting trajectories we see that PPO learns to run more quickly at the cost of falling over and incurring penalties}} 
    \label{fig:obstacles_halfcheetah}
\end{figure}

We now, turn to another transfer learning task: controlling the Halfcheetah robot in the task locomotion with obstacles (described in Section \ref{sec:benchmark_ecorobot}). 
In Figure \ref{fig:obstacles_halfcheetah}, we observe that the results are qualitatively similar to the maze with stepping-stones.
PPO performs much better in this task, however, arguably to its capability of handling complex continuous-control tasks.
As we show in visualizations of the trajectories in our companion website, however, policies learned by PPO are fast but ''bumpy'', incurring the control cost penalty of this robot and often flipping it over, indicating that PPO has not learned to adjust its behavior to obstacles of varying size but to ignore them. 

\begin{figure}
    \centering
        \includegraphics[width=0.9\linewidth]{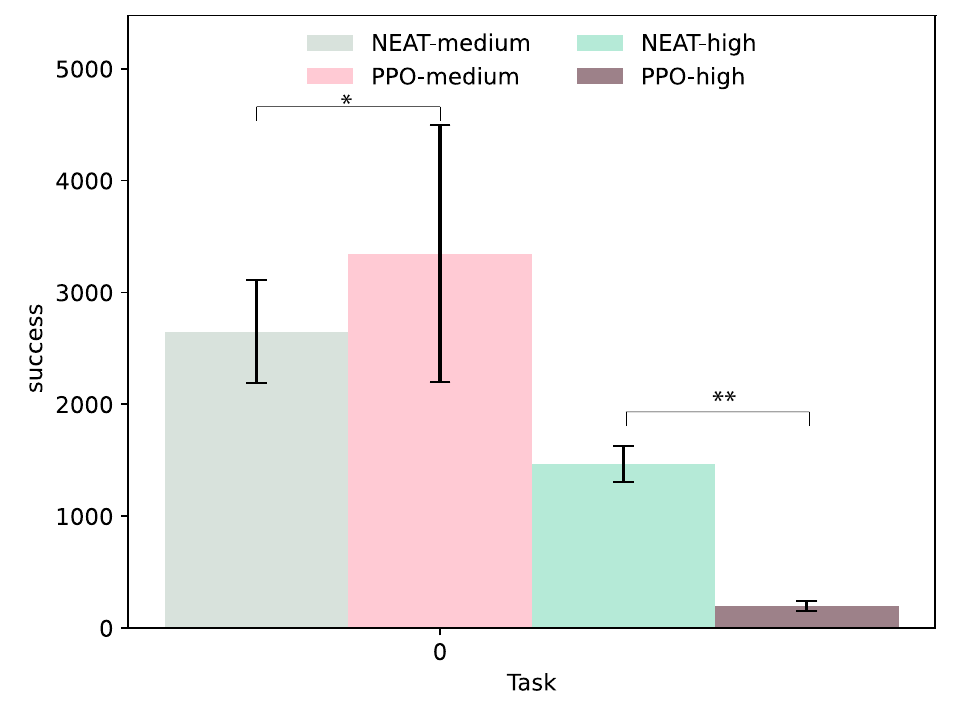}
    \caption{Ablation for the task locomotion with obstacles:{ \normalfont we consider tasks with medium and high level of difficulty where all obstacles have the same height} }
    \label{fig:obstacles_nocurric}
\end{figure}

In a similar spirit to our ablation for the N-parity task, we now investigate how our observations change when we disable the curriculum in the task locomotion with obstacles.
We consider tasks where all obstacles have the same height and study two levels of difficulty: in the medium-difficulty task all obstacles have the height of the second level of the original task while in the high-difficulty task they have the height of the third level.
Figure \ref{fig:obstacles_nocurric} presents the performance of NEAT and PPO for the two cases.
We observe that, for the medium-difficulty task, NEAT's performance drops while the performance of PPO remains unchanged.
Thus, similarly to our N-parity task we see that NEAT has benefited from a curriculum.
For the high-difficulty task, however, PPO fails dramatically while NEAT's performance degrades more gracefully.
Inspecting trajectories shows that PPO has gotten stuck at the first obstacle while NEAT overpasses the obstacles yet primarily by flipping over.
Thus, the absence of a curriculum renders the final level unsolvable.
Note that this experiment does not show that PPO benefited from a curriculum despite its increased performance compared to the difficult setting: as the behavioral trajectories indicate it did not master the final level but progressed by flipping over.

Our study of transfer learning in ecorobot has led to very similar qualitative conclusions to the ones we saw in the stepping gates benchmark.
We view this as an encouraging indication that the two benchmarks capture the notion of transfer learning we were interested in.
A difference is that CMA-ES performs worse than NEAT.
While it is hard to reason why this is the case without more experimentation, we suspect that CMA-ES can benefit from further tuning.
In its ability to grow its own architecture, NEAT can prove more robust (we noticed that NEAT grows large networks for the stepping gates tasks while the networks found for this maze are surprisingly small).
%Note that despite large differences in the means our statistical tests did not indicate significance due to the large differences in the variances of methods.

\subsubsection{Morphology complexification}
\label{sec:results_complex_morphology}

\begin{figure}
    \centering
    \includegraphics[width=0.9\linewidth]{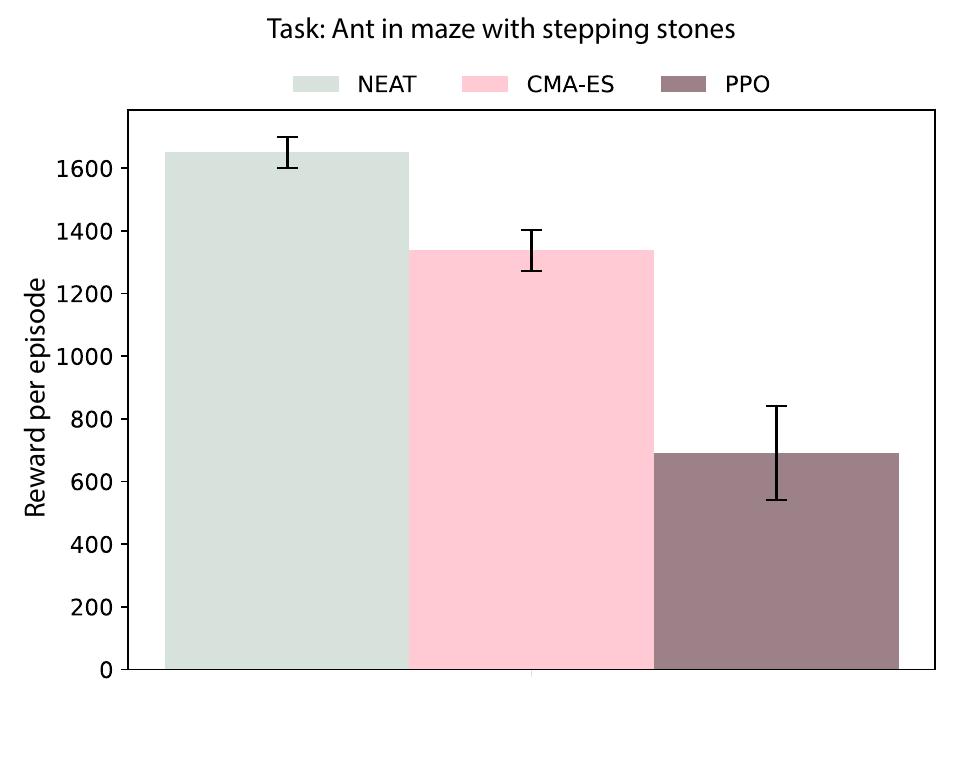}
    \caption{Effect of complexifying the robot morphology { \normalfont NEAT can no longer progress in the maze with stepping stones (contrast with Figure \ref{fig:stepping_rob})}}
    \label{fig:complex_morpho}
\end{figure}

So far our experiments have demonstrated that, among the investigated algorithms, NEAT is a promising for transfer learning.
Yet, ultimately, we are in need of algorithms that can not only transfer skills but also continue to do so as the problem complexity scales up.
Our next experiment will show that this ability does not readily arise with NEAT.
To scale up the complexity of the task without changing the nature of the challenge, we replace SimpleRob with the ant robot in the stepping stones maze.
Figure \ref{fig:complex_morpho} shows that, this time, the robot controlled by NEAT reaches only the first stepping stone in the maze.

\subsubsection{Avoidance of local optima}\label{sec:results_indirect}
\begin{figure}
    \centering
    \includegraphics[width=0.9\linewidth]{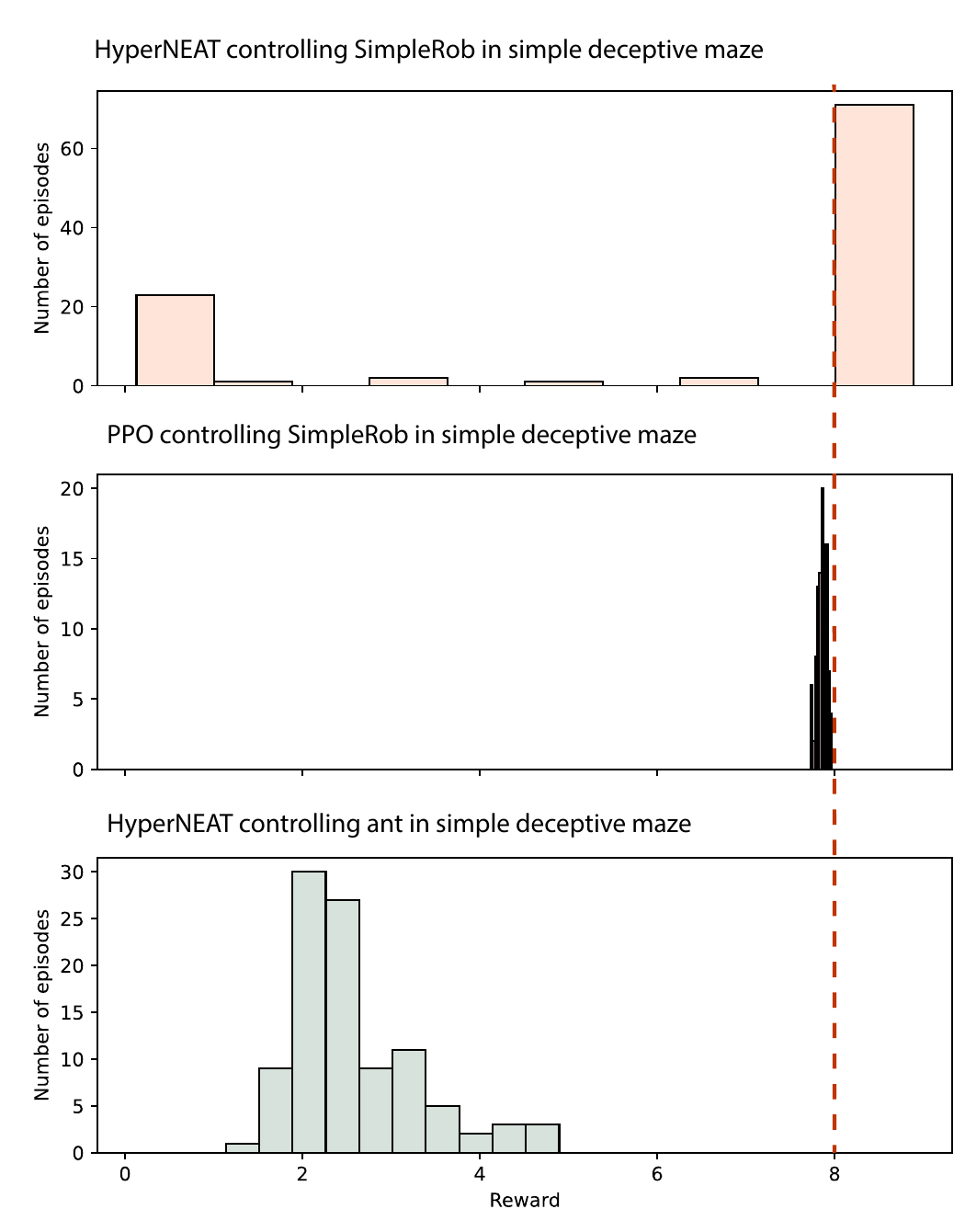}
    \caption{Histogram of rewards for multiple evaluation episodes of a single trained policy. Red dashed line indicates the reward value indicated that the episode was successful: {\normalfont (top) HyperNEAT escapes the local optimum when controlling SimpleRob
    (middle) PPO gets stuck on the deceptive wall, a policy that incurs close to optimal reward.
    (bottom)
    When replacing SimpleRob with the ant, HyperNEAT can no longer solve the task.} 
    }
    \label{fig:deceptive}
\end{figure}

Our experiments have painted a rather grim picture for indirect encodings, as we've seen HyperNEAT consistently failing in our transfer learning tasks.
Here, we would like to show that there are cases where an indirect encoding can be useful in its ability to perform large jumps in phenotypic space: tasks with local optima.
Figure \ref{fig:deceptive} presents evidence for such a case: in a simplified version of the deceptive maze (illustrated on the right of Figure \ref{fig:ecorobot_tasks}), HyperNEAT is able to avoid the local optimum and reach the target while PPO remains stuck at the nearby wall. 
Resonating with the results in the previous section, however, this ability of HyperNEAT disappears when we replace the simple morphology with the ant robot.
It also disappears when we consider the larger deceptive maze with multiple traps where methods that explicitly optimize for diversity excel\citep{chalumeau_neuroevolution_2023, mouret2015illuminating}, pointing to the hypothesis that, when it comes to local optima tasks, indirect maps can outcompete RL but lose to NE methods that explicitly preserve diversity.

\section{Discussion}
We have introduced two benchmarks for evaluating the transfer learning abilities of agents and evaluated different NE and RL methods.
Our experiments reproduce some known results from the NE literature, add some new observations, and demonstrate that the benchmarks reveal meaningful differences among approaches.
Our main conclusion is that direct encodings fair better in transfer learning, with architecture search (through NEAT) performing in general on par with CMA-ES.
However none of the considered methods manages to address both complex locomotion and transfer learning. We believe that, considering the encouraging performance of NEAT and the benefits of architecture search in general, future work should attempt to address its limitations.
A thought-provoking question is whether one can bring the benefits of indirect encodings (large evolutionary leaps, avoidance of local optima) and the benefits of direct encodings (transfer learning, as indicated here) to design encodings that can accelerate evolution.
Such questions are prominent in the study of biological evolution and, in particular, evo-devo, where hypotheses point to the fact that evolution can be significantly sped up when development supports modular and hierarchical morphologies and neural architectures \citep{cisek_resynthesizing_2019,knoll_early_1999}.

Our empirical investigation was limited in various respects.
First, RL is a rich, ever-expanding family of algorithms, some of which are designed to address transfer learning and would potentially perform well in our benchmarks.
Our objective, however, was not to outcompete state-of-the-art RL but to show that some NE algorithms are inherently better suited for solving transfer learning tasks than others.
Second, we have chosen a simple instantiation of a Quality-Diversity algorithm while there is a variety of choices, such as employing RL or an indirect encoding.
While this could bring some benefits, our analysis has offered the intriguing observation that transfer learning tasks are fundamentally different from deceptive tasks, suggesting that optimizing for diversity is not sufficient.
Finally, despite our effort to include a wide diversity of NE algorithms, we expect that richer conclusions can be reached through a large-scale analysis that performs more extensive hyperparameter tuning.
Finally, we plan to enrich our robotic benchmark, ecorobot, with additional features , such as camera observations, multi-robot scenarios, and more complex item interactions to support the study of more complex skills.

% explain that there variants in RL that we expect to solve these tasks. also the recent version of map-elites

%Todo: mention generative models for designing environments and that we still need structured task generation. also mention that we continue to support with images. 

% limitations of the study: although we have tried to benchmark RL we still need to study more sophisticated versions. however our algorithms were very robust. also map-elites is still developing

% we believe that our environments have shown many interesting insights. the stepping stones maze is a smlight variation of the maze and our study shows it shows a meaning difference. the obstacles are also interesting. the ability to easily switch robots enables us to disentangle the effect of complex locomotion from the cognitive challenge 

%\begin{acks}
%To Robert, for the bagels and explaining CMYK and color spaces.
%\end{acks}

\section*{Acknowledgments}

Funded by the European Union (ERC, GROW-AI, 101045094). Views and opinions expressed are however those of the authors only and do not necessarily reflect those of the European Union or the European Research Council. Neither the European Union nor the granting authority can be held responsible for them.

\bibliographystyle{ACM-Reference-Format}
\bibliography{sample-base}


\begin{thebibliography}{50}

%%% the Web address.

\ifx \showCODEN    \undefined \def \showCODEN     #1{\unskip}     \fi
\ifx \showDOI      \undefined \def \showDOI       #1{#1}\fi
\ifx \showISBNx    \undefined \def \showISBNx     #1{\unskip}     \fi
\ifx \showISBNxiii \undefined \def \showISBNxiii  #1{\unskip}     \fi
\ifx \showISSN     \undefined \def \showISSN      #1{\unskip}     \fi
\ifx \showLCCN     \undefined \def \showLCCN      #1{\unskip}     \fi
\ifx \shownote     \undefined \def \shownote      #1{#1}          \fi
\ifx \showarticletitle \undefined \def \showarticletitle #1{#1}   \fi
\ifx \showURL      \undefined \def \showURL       {\relax}        \fi
% invisible to TeX
\providecommand\bibfield[2]{#2}
\providecommand\bibinfo[2]{#2}
\providecommand\natexlab[1]{#1}
\providecommand\showeprint[2][]{arXiv:#2}

\bibitem[Bengio et~al\mbox{.}(2009)]%
        {bengio2009curriculum}
\bibfield{author}{\bibinfo{person}{Yoshua Bengio}, \bibinfo{person}{J{\'e}r{\^o}me Louradour}, \bibinfo{person}{Ronan Collobert}, {and} \bibinfo{person}{Jason Weston}.} \bibinfo{year}{2009}\natexlab{}.
\newblock \showarticletitle{Curriculum learning}. In \bibinfo{booktitle}{\emph{Proceedings of the 26th annual international conference on machine learning}}. \bibinfo{pages}{41--48}.
\newblock


\bibitem[Chalumeau et~al\mbox{.}(2023)]%
        {chalumeau_neuroevolution_2023}
\bibfield{author}{\bibinfo{person}{Felix Chalumeau}, \bibinfo{person}{Raphael Boige}, \bibinfo{person}{Bryan Lim}, \bibinfo{person}{Valentin Macé}, \bibinfo{person}{Maxime Allard}, \bibinfo{person}{Arthur Flajolet}, \bibinfo{person}{Antoine Cully}, {and} \bibinfo{person}{Thomas Pierrot}.} \bibinfo{year}{2023}\natexlab{}.
\newblock \bibinfo{title}{Neuroevolution is a {Competitive} {Alternative} to {Reinforcement} {Learning} for {Skill} {Discovery}}.
\newblock
\newblock
\urldef\tempurl%
\url{https://doi.org/10.48550/arXiv.2210.03516}
\showDOI{\tempurl}
\newblock
\shownote{arXiv:2210.03516 [cs]}.


\bibitem[Chalumeau et~al\mbox{.}(2024)]%
        {chalumeau2024qdax}
\bibfield{author}{\bibinfo{person}{Felix Chalumeau}, \bibinfo{person}{Bryan Lim}, \bibinfo{person}{Raphael Boige}, \bibinfo{person}{Maxime Allard}, \bibinfo{person}{Luca Grillotti}, \bibinfo{person}{Manon Flageat}, \bibinfo{person}{Valentin Mac{\'e}}, \bibinfo{person}{Guillaume Richard}, \bibinfo{person}{Arthur Flajolet}, \bibinfo{person}{Thomas Pierrot}, {et~al\mbox{.}}} \bibinfo{year}{2024}\natexlab{}.
\newblock \showarticletitle{Qdax: A library for quality-diversity and population-based algorithms with hardware acceleration}.
\newblock \bibinfo{journal}{\emph{Journal of Machine Learning Research}} \bibinfo{volume}{25}, \bibinfo{number}{108} (\bibinfo{year}{2024}), \bibinfo{pages}{1--16}.
\newblock


\bibitem[Cisek(2019)]%
        {cisek_resynthesizing_2019}
\bibfield{author}{\bibinfo{person}{Paul Cisek}.} \bibinfo{year}{2019}\natexlab{}.
\newblock \showarticletitle{Resynthesizing behavior through phylogenetic refinement}.
\newblock \bibinfo{journal}{\emph{Attention, Perception, \& Psychophysics}} \bibinfo{volume}{81}, \bibinfo{number}{7} (\bibinfo{date}{Oct.} \bibinfo{year}{2019}), \bibinfo{pages}{2265--2287}.
\newblock
\showISSN{1943-393X}
\urldef\tempurl%
\url{https://doi.org/10.3758/s13414-019-01760-1}
\showDOI{\tempurl}


\bibitem[Colas et~al\mbox{.}(2022)]%
        {colas_autotelic_2022}
\bibfield{author}{\bibinfo{person}{Cédric Colas}, \bibinfo{person}{Tristan Karch}, \bibinfo{person}{Olivier Sigaud}, {and} \bibinfo{person}{Pierre-Yves Oudeyer}.} \bibinfo{year}{2022}\natexlab{}.
\newblock \bibinfo{title}{Autotelic {Agents} with {Intrinsically} {Motivated} {Goal}-{Conditioned} {Reinforcement} {Learning}: a {Short} {Survey}}.
\newblock
\newblock
\urldef\tempurl%
\url{https://doi.org/10.48550/arXiv.2012.09830}
\showDOI{\tempurl}
\newblock
\shownote{arXiv:2012.09830 [cs]}.


\bibitem[Crosby et~al\mbox{.}(2019)]%
        {crosby_animal-ai_2019}
\bibfield{author}{\bibinfo{person}{Matthew Crosby}, \bibinfo{person}{Benjamin Beyret}, {and} \bibinfo{person}{Marta Halina}.} \bibinfo{year}{2019}\natexlab{}.
\newblock \showarticletitle{The {Animal}-{AI} {Olympics}}.
\newblock \bibinfo{journal}{\emph{Nature Machine Intelligence}} \bibinfo{volume}{1}, \bibinfo{number}{5} (\bibinfo{date}{May} \bibinfo{year}{2019}), \bibinfo{pages}{257--257}.
\newblock
\showISSN{2522-5839}
\urldef\tempurl%
\url{https://doi.org/10.1038/s42256-019-0050-3}
\showDOI{\tempurl}
\newblock
\shownote{Publisher: Nature Publishing Group}.


\bibitem[Ecoffet et~al\mbox{.}(2021)]%
        {ecoffet_first_2021}
\bibfield{author}{\bibinfo{person}{Adrien Ecoffet}, \bibinfo{person}{Joost Huizinga}, \bibinfo{person}{Joel Lehman}, \bibinfo{person}{Kenneth~O. Stanley}, {and} \bibinfo{person}{Jeff Clune}.} \bibinfo{year}{2021}\natexlab{}.
\newblock \showarticletitle{First return, then explore}.
\newblock \bibinfo{journal}{\emph{Nature}} \bibinfo{volume}{590}, \bibinfo{number}{7847} (\bibinfo{date}{Feb.} \bibinfo{year}{2021}), \bibinfo{pages}{580--586}.
\newblock
\showISSN{1476-4687}
\urldef\tempurl%
\url{https://doi.org/10.1038/s41586-020-03157-9}
\showDOI{\tempurl}
\newblock
\shownote{Publisher: Nature Publishing Group}.


\bibitem[Ellefsen et~al\mbox{.}(2015)]%
        {ellefsen_neural_2015}
\bibfield{author}{\bibinfo{person}{Kai~Olav Ellefsen}, \bibinfo{person}{Jean-Baptiste Mouret}, {and} \bibinfo{person}{Jeff Clune}.} \bibinfo{year}{2015}\natexlab{}.
\newblock \showarticletitle{Neural {Modularity} {Helps} {Organisms} {Evolve} to {Learn} {New} {Skills} without {Forgetting} {Old} {Skills}}.
\newblock \bibinfo{journal}{\emph{PLOS Computational Biology}} \bibinfo{volume}{11}, \bibinfo{number}{4} (\bibinfo{date}{April} \bibinfo{year}{2015}), \bibinfo{pages}{e1004128}.
\newblock
\showISSN{1553-7358}
\urldef\tempurl%
\url{https://doi.org/10.1371/journal.pcbi.1004128}
\showDOI{\tempurl}


\bibitem[Eysenbach et~al\mbox{.}(2018)]%
        {eysenbach_diversity_2018}
\bibfield{author}{\bibinfo{person}{Benjamin Eysenbach}, \bibinfo{person}{Abhishek Gupta}, \bibinfo{person}{Julian Ibarz}, {and} \bibinfo{person}{Sergey Levine}.} \bibinfo{year}{2018}\natexlab{}.
\newblock \bibinfo{title}{Diversity is {All} {You} {Need}: {Learning} {Skills} without a {Reward} {Function}}.
\newblock
\newblock
\urldef\tempurl%
\url{https://doi.org/10.48550/arXiv.1802.06070}
\showDOI{\tempurl}
\newblock
\shownote{arXiv:1802.06070 [cs]}.


\bibitem[Freeman et~al\mbox{.}(2021)]%
        {freeman_brax_2021}
\bibfield{author}{\bibinfo{person}{C.~Daniel Freeman}, \bibinfo{person}{Erik Frey}, \bibinfo{person}{Anton Raichuk}, \bibinfo{person}{Sertan Girgin}, \bibinfo{person}{Igor Mordatch}, {and} \bibinfo{person}{Olivier Bachem}.} \bibinfo{year}{2021}\natexlab{}.
\newblock \bibinfo{title}{Brax -- {A} {Differentiable} {Physics} {Engine} for {Large} {Scale} {Rigid} {Body} {Simulation}}.
\newblock
\newblock
\urldef\tempurl%
\url{https://doi.org/10.48550/arXiv.2106.13281}
\showDOI{\tempurl}
\newblock
\shownote{arXiv:2106.13281 [cs]}.


\bibitem[Greenbury et~al\mbox{.}(2022)]%
        {greenbury_structure_2022}
\bibfield{author}{\bibinfo{person}{Sam~F. Greenbury}, \bibinfo{person}{Ard~A. Louis}, {and} \bibinfo{person}{Sebastian~E. Ahnert}.} \bibinfo{year}{2022}\natexlab{}.
\newblock \showarticletitle{The structure of genotype-phenotype maps makes fitness landscapes navigable}.
\newblock \bibinfo{journal}{\emph{Nature Ecology \& Evolution}} \bibinfo{volume}{6}, \bibinfo{number}{11} (\bibinfo{date}{Nov.} \bibinfo{year}{2022}), \bibinfo{pages}{1742--1752}.
\newblock
\showISSN{2397-334X}
\urldef\tempurl%
\url{https://doi.org/10.1038/s41559-022-01867-z}
\showDOI{\tempurl}
\newblock
\shownote{Publisher: Nature Publishing Group}.


\bibitem[Grillotti and Cully(2023)]%
        {grillotti_kheperax_2023}
\bibfield{author}{\bibinfo{person}{Luca Grillotti} {and} \bibinfo{person}{Antoine Cully}.} \bibinfo{year}{2023}\natexlab{}.
\newblock \showarticletitle{Kheperax: a {Lightweight} {JAX}-based {Robot} {Control} {Environment} for {Benchmarking} {Quality}-{Diversity} {Algorithms}}. In \bibinfo{booktitle}{\emph{Proceedings of the {Companion} {Conference} on {Genetic} and {Evolutionary} {Computation}}} \emph{(\bibinfo{series}{{GECCO} '23 {Companion}})}. \bibinfo{publisher}{Association for Computing Machinery}, \bibinfo{address}{New York, NY, USA}, \bibinfo{pages}{2163--2165}.
\newblock
\showISBNx{9798400701207}
\urldef\tempurl%
\url{https://doi.org/10.1145/3583133.3596387}
\showDOI{\tempurl}


\bibitem[Gruau(1992)]%
        {gruau_genetic_1992}
\bibfield{author}{\bibinfo{person}{F. Gruau}.} \bibinfo{year}{1992}\natexlab{}.
\newblock \showarticletitle{Genetic synthesis of {Boolean} neural networks with a cell rewriting developmental process}. In \bibinfo{booktitle}{\emph{[{Proceedings}] {COGANN}-92: {International} {Workshop} on {Combinations} of {Genetic} {Algorithms} and {Neural} {Networks}}}. \bibinfo{publisher}{IEEE Comput. Soc. Press}, \bibinfo{address}{Baltimore, MD, USA}, \bibinfo{pages}{55--74}.
\newblock
\showISBNx{978-0-8186-2787-3}
\urldef\tempurl%
\url{https://doi.org/10.1109/COGANN.1992.273948}
\showDOI{\tempurl}


\bibitem[Gruau(1994)]%
        {gruau_automatic_1994}
\bibfield{author}{\bibinfo{person}{Frédéric Gruau}.} \bibinfo{year}{1994}\natexlab{}.
\newblock \showarticletitle{Automatic {Definition} of {Modular} {Neural} {Networks}}.
\newblock \bibinfo{journal}{\emph{Adaptive Behavior}} \bibinfo{volume}{3}, \bibinfo{number}{2} (\bibinfo{date}{Sept.} \bibinfo{year}{1994}), \bibinfo{pages}{151--183}.
\newblock
\showISSN{1059-7123, 1741-2633}
\urldef\tempurl%
\url{https://doi.org/10.1177/105971239400300202}
\showDOI{\tempurl}


\bibitem[Hansen(2023)]%
        {hansen_cma_2023}
\bibfield{author}{\bibinfo{person}{Nikolaus Hansen}.} \bibinfo{year}{2023}\natexlab{}.
\newblock \bibinfo{title}{The {CMA} {Evolution} {Strategy}: {A} {Tutorial}}.
\newblock
\newblock
\urldef\tempurl%
\url{https://doi.org/10.48550/arXiv.1604.00772}
\showDOI{\tempurl}
\newblock
\shownote{arXiv:1604.00772 [cs]}.


\bibitem[Huang et~al\mbox{.}(2021)]%
        {huang_continual_2021}
\bibfield{author}{\bibinfo{person}{Yizhou Huang}, \bibinfo{person}{Kevin Xie}, \bibinfo{person}{Homanga Bharadhwaj}, {and} \bibinfo{person}{Florian Shkurti}.} \bibinfo{year}{2021}\natexlab{}.
\newblock \bibinfo{title}{Continual {Model}-{Based} {Reinforcement} {Learning} with {Hypernetworks}}.
\newblock
\newblock
\urldef\tempurl%
\url{https://doi.org/10.48550/arXiv.2009.11997}
\showDOI{\tempurl}
\newblock
\shownote{arXiv:2009.11997 [cs]}.


\bibitem[Kaelbling(1993)]%
        {kaelbling_learning_1993}
\bibfield{author}{\bibinfo{person}{Leslie~Pack Kaelbling}.} \bibinfo{year}{1993}\natexlab{}.
\newblock \showarticletitle{Learning to achieve goals}. In \bibinfo{booktitle}{\emph{{IJCAI}}}, Vol.~\bibinfo{volume}{2}. \bibinfo{publisher}{Citeseer}, \bibinfo{pages}{1094--1098}.
\newblock
\urldef\tempurl%
\url{https://citeseerx.ist.psu.edu/document?repid=rep1&type=pdf&doi=6df43f70f383007a946448122b75918e3a9d6682}
\showURL{%
\tempurl}


\bibitem[Khetarpal et~al\mbox{.}(2022)]%
        {khetarpal_towards_2022}
\bibfield{author}{\bibinfo{person}{Khimya Khetarpal}, \bibinfo{person}{Matthew Riemer}, \bibinfo{person}{Irina Rish}, {and} \bibinfo{person}{Doina Precup}.} \bibinfo{year}{2022}\natexlab{}.
\newblock \bibinfo{title}{Towards {Continual} {Reinforcement} {Learning}: {A} {Review} and {Perspectives}}.
\newblock
\newblock
\urldef\tempurl%
\url{https://doi.org/10.48550/arXiv.2012.13490}
\showDOI{\tempurl}
\newblock
\shownote{arXiv:2012.13490 [cs]}.


\bibitem[Kingma and Ba(2017)]%
        {kingma_adam_2017}
\bibfield{author}{\bibinfo{person}{Diederik~P. Kingma} {and} \bibinfo{person}{Jimmy Ba}.} \bibinfo{year}{2017}\natexlab{}.
\newblock \bibinfo{title}{Adam: {A} {Method} for {Stochastic} {Optimization}}.
\newblock
\newblock
\urldef\tempurl%
\url{https://doi.org/10.48550/arXiv.1412.6980}
\showDOI{\tempurl}
\newblock
\shownote{arXiv:1412.6980 [cs]}.


\bibitem[Knoll and Carroll(1999)]%
        {knoll_early_1999}
\bibfield{author}{\bibinfo{person}{Andrew~H. Knoll} {and} \bibinfo{person}{Sean~B. Carroll}.} \bibinfo{year}{1999}\natexlab{}.
\newblock \showarticletitle{Early {Animal} {Evolution}: {Emerging} {Views} from {Comparative} {Biology} and {Geology}}.
\newblock \bibinfo{journal}{\emph{Science}} \bibinfo{volume}{284}, \bibinfo{number}{5423} (\bibinfo{date}{June} \bibinfo{year}{1999}), \bibinfo{pages}{2129--2137}.
\newblock
\showISSN{0036-8075, 1095-9203}
\urldef\tempurl%
\url{https://doi.org/10.1126/science.284.5423.2129}
\showDOI{\tempurl}


\bibitem[Koza and Rice(1991)]%
        {koza_genetic_1991}
\bibfield{author}{\bibinfo{person}{J.R. Koza} {and} \bibinfo{person}{J.P. Rice}.} \bibinfo{year}{1991}\natexlab{}.
\newblock \showarticletitle{Genetic generation of both the weights and architecture for a neural network}. In \bibinfo{booktitle}{\emph{{IJCNN}-91-{Seattle} {International} {Joint} {Conference} on {Neural} {Networks}}}, Vol.~\bibinfo{volume}{ii}. \bibinfo{publisher}{IEEE}, \bibinfo{address}{Seattle, WA, USA}, \bibinfo{pages}{397--404}.
\newblock
\showISBNx{978-0-7803-0164-1}
\urldef\tempurl%
\url{https://doi.org/10.1109/IJCNN.1991.155366}
\showDOI{\tempurl}


\bibitem[Koza and Rice(1992)]%
        {koza_genetic}
\bibfield{author}{\bibinfo{person}{J.R. Koza} {and} \bibinfo{person}{J.P. Rice}.} \bibinfo{year}{1992}\natexlab{}.
\newblock \bibinfo{title}{Genetic {Programming}}.
\newblock
\newblock
\urldef\tempurl%
\url{https://mitpress.mit.edu/9780262527910/genetic-programming/}
\showURL{%
\tempurl}


\bibitem[Koza({[n.\,d.]})]%
        {koza_genetic_nodate}
\bibfield{author}{\bibinfo{person}{John~R Koza}.} \bibinfo{year}{[n.\,d.]}\natexlab{}.
\newblock \showarticletitle{{genetic} {programming}: {a} {paradigm} {for} {genetically} {breeding} {populations} {of} {computer} {programs} {to} {solve} {problems}}.
\newblock  (\bibinfo{year}{[n.\,d.]}).
\newblock


\bibitem[Lange(2022)]%
        {evosax2022github}
\bibfield{author}{\bibinfo{person}{Robert~Tjarko Lange}.} \bibinfo{year}{2022}\natexlab{}.
\newblock \showarticletitle{evosax: JAX-based Evolution Strategies}.
\newblock \bibinfo{journal}{\emph{arXiv preprint arXiv:2212.04180}} (\bibinfo{year}{2022}).
\newblock


\bibitem[Lehman and Stanley(2011)]%
        {lehman_abandoning_2011}
\bibfield{author}{\bibinfo{person}{Joel Lehman} {and} \bibinfo{person}{Kenneth~O. Stanley}.} \bibinfo{year}{2011}\natexlab{}.
\newblock \showarticletitle{Abandoning {Objectives}: {Evolution} {Through} the {Search} for {Novelty} {Alone}}.
\newblock \bibinfo{journal}{\emph{Evolutionary Computation}} \bibinfo{volume}{19}, \bibinfo{number}{2} (\bibinfo{date}{June} \bibinfo{year}{2011}), \bibinfo{pages}{189--223}.
\newblock
\showISSN{1063-6560}
\urldef\tempurl%
\url{https://doi.org/10.1162/EVCO_a_00025}
\showDOI{\tempurl}


\bibitem[Lim et~al\mbox{.}(2022)]%
        {lim_accelerated_2022}
\bibfield{author}{\bibinfo{person}{Bryan Lim}, \bibinfo{person}{Maxime Allard}, \bibinfo{person}{Luca Grillotti}, {and} \bibinfo{person}{Antoine Cully}.} \bibinfo{year}{2022}\natexlab{}.
\newblock \bibinfo{title}{Accelerated {Quality}-{Diversity} through {Massive} {Parallelism}}.
\newblock
\newblock
\urldef\tempurl%
\url{https://doi.org/10.48550/arXiv.2202.01258}
\showDOI{\tempurl}
\newblock
\shownote{arXiv:2202.01258 [cs]}.


\bibitem[Lucas and Prémont-Schwarz(2024)]%
        {lucas_articulated_2024}
\bibfield{author}{\bibinfo{person}{Jeremy Lucas} {and} \bibinfo{person}{Isabeau Prémont-Schwarz}.} \bibinfo{year}{2024}\natexlab{}.
\newblock \bibinfo{title}{Articulated {Animal} {AI}: {An} {Environment} for {Animal}-like {Cognition} in a {Limbed} {Agent}}.
\newblock
\newblock
\urldef\tempurl%
\url{https://doi.org/10.48550/arXiv.2410.09275}
\showDOI{\tempurl}
\newblock
\shownote{arXiv:2410.09275 [cs]}.


\bibitem[Matthews et~al\mbox{.}(2024)]%
        {matthews_kinetix_2024}
\bibfield{author}{\bibinfo{person}{Michael Matthews}, \bibinfo{person}{Michael Beukman}, \bibinfo{person}{Chris Lu}, {and} \bibinfo{person}{Jakob Foerster}.} \bibinfo{year}{2024}\natexlab{}.
\newblock \bibinfo{title}{Kinetix: {Investigating} the {Training} of {General} {Agents} through {Open}-{Ended} {Physics}-{Based} {Control} {Tasks}}.
\newblock
\newblock
\urldef\tempurl%
\url{https://doi.org/10.48550/arXiv.2410.23208}
\showDOI{\tempurl}
\newblock
\shownote{arXiv:2410.23208 [cs]}.


\bibitem[Moriarty and Miikkulainen(1996)]%
        {moriarty_efficient_1996}
\bibfield{author}{\bibinfo{person}{David~E. Moriarty} {and} \bibinfo{person}{Risto Miikkulainen}.} \bibinfo{year}{1996}\natexlab{}.
\newblock \showarticletitle{Efficient reinforcement learning through symbiotic evolution}.
\newblock \bibinfo{journal}{\emph{Machine Learning}} \bibinfo{volume}{22}, \bibinfo{number}{1} (\bibinfo{date}{March} \bibinfo{year}{1996}), \bibinfo{pages}{11--32}.
\newblock
\showISSN{1573-0565}
\urldef\tempurl%
\url{https://doi.org/10.1007/BF00114722}
\showDOI{\tempurl}


\bibitem[Mouret and Clune(2015)]%
        {mouret2015illuminating}
\bibfield{author}{\bibinfo{person}{Jean-Baptiste Mouret} {and} \bibinfo{person}{Jeff Clune}.} \bibinfo{year}{2015}\natexlab{}.
\newblock \showarticletitle{Illuminating search spaces by mapping elites}.
\newblock \bibinfo{journal}{\emph{arXiv preprint arXiv:1504.04909}} (\bibinfo{year}{2015}).
\newblock


\bibitem[Najarro et~al\mbox{.}(2022)]%
        {najarro_hypernca_2022}
\bibfield{author}{\bibinfo{person}{Elias Najarro}, \bibinfo{person}{Shyam Sudhakaran}, \bibinfo{person}{Claire Glanois}, {and} \bibinfo{person}{Sebastian Risi}.} \bibinfo{year}{2022}\natexlab{}.
\newblock \bibinfo{title}{{HyperNCA}: {Growing} {Developmental} {Networks} with {Neural} {Cellular} {Automata}}.
\newblock
\newblock
\urldef\tempurl%
\url{http://arxiv.org/abs/2204.11674}
\showURL{%
\tempurl}
\newblock
\shownote{arXiv:2204.11674 [cs]}.


\bibitem[Nisioti et~al\mbox{.}(2024)]%
        {nisioti_growing_2024}
\bibfield{author}{\bibinfo{person}{Eleni Nisioti}, \bibinfo{person}{Erwan Plantec}, \bibinfo{person}{Milton Montero}, \bibinfo{person}{Joachim Pedersen}, {and} \bibinfo{person}{Sebastian Risi}.} \bibinfo{year}{2024}\natexlab{}.
\newblock \showarticletitle{Growing {Artificial} {Neural} {Networks} for {Control}: the {Role} of {Neuronal} {Diversity}}. In \bibinfo{booktitle}{\emph{Proceedings of the {Genetic} and {Evolutionary} {Computation} {Conference} {Companion}}}. \bibinfo{publisher}{ACM}, \bibinfo{address}{Melbourne VIC Australia}, \bibinfo{pages}{175--178}.
\newblock
\showISBNx{9798400704956}
\urldef\tempurl%
\url{https://doi.org/10.1145/3638530.3654356}
\showDOI{\tempurl}


\bibitem[Papavasileiou et~al\mbox{.}(2021)]%
        {papavasileiou_systematic_2021}
\bibfield{author}{\bibinfo{person}{Evgenia Papavasileiou}, \bibinfo{person}{Jan Cornelis}, {and} \bibinfo{person}{Bart Jansen}.} \bibinfo{year}{2021}\natexlab{}.
\newblock \showarticletitle{A {Systematic} {Literature} {Review} of the {Successors} of “{NeuroEvolution} of {Augmenting} {Topologies}”}.
\newblock \bibinfo{journal}{\emph{Evolutionary Computation}} \bibinfo{volume}{29}, \bibinfo{number}{1} (\bibinfo{date}{March} \bibinfo{year}{2021}), \bibinfo{pages}{1--73}.
\newblock
\showISSN{1063-6560}
\urldef\tempurl%
\url{https://doi.org/10.1162/evco_a_00282}
\showDOI{\tempurl}


\bibitem[Pugh et~al\mbox{.}(2016)]%
        {pugh_quality_2016}
\bibfield{author}{\bibinfo{person}{Justin~K. Pugh}, \bibinfo{person}{Lisa~B. Soros}, {and} \bibinfo{person}{Kenneth~O. Stanley}.} \bibinfo{year}{2016}\natexlab{}.
\newblock \showarticletitle{Quality {Diversity}: {A} {New} {Frontier} for {Evolutionary} {Computation}}.
\newblock \bibinfo{journal}{\emph{Frontiers in Robotics and AI}}  \bibinfo{volume}{3} (\bibinfo{date}{July} \bibinfo{year}{2016}).
\newblock
\showISSN{2296-9144}
\urldef\tempurl%
\url{https://doi.org/10.3389/frobt.2016.00040}
\showDOI{\tempurl}
\newblock
\shownote{Publisher: Frontiers}.


\bibitem[Risi et~al\mbox{.}(2025)]%
        {risi:book25}
\bibfield{author}{\bibinfo{person}{Sebastian Risi}, \bibinfo{person}{David Ha}, \bibinfo{person}{Yujin Tang}, {and} \bibinfo{person}{Risto Miikkulainen}.} \bibinfo{year}{2025}\natexlab{}.
\newblock \bibinfo{booktitle}{\emph{Neuroevolution: Harnessing Creativity in AI Model Design}}.
\newblock \bibinfo{publisher}{MIT Press}, \bibinfo{address}{Cambridge, MA}.
\newblock
\urldef\tempurl%
\url{http://www.cs.utexas.edu/users/ai-lab?risi:book25}
\showURL{%
\tempurl}


\bibitem[Risi et~al\mbox{.}(2010)]%
        {risi_evolving_2010}
\bibfield{author}{\bibinfo{person}{Sebastian Risi}, \bibinfo{person}{Joel Lehman}, {and} \bibinfo{person}{Kenneth~O. Stanley}.} \bibinfo{year}{2010}\natexlab{}.
\newblock \showarticletitle{Evolving the placement and density of neurons in the hyperneat substrate}. In \bibinfo{booktitle}{\emph{Proceedings of the 12th annual conference on {Genetic} and evolutionary computation}} \emph{(\bibinfo{series}{{GECCO} '10})}. \bibinfo{publisher}{Association for Computing Machinery}, \bibinfo{address}{New York, NY, USA}, \bibinfo{pages}{563--570}.
\newblock
\showISBNx{978-1-4503-0072-8}
\urldef\tempurl%
\url{https://doi.org/10.1145/1830483.1830589}
\showDOI{\tempurl}


\bibitem[Salimans et~al\mbox{.}(2017)]%
        {salimans_evolution_2017}
\bibfield{author}{\bibinfo{person}{Tim Salimans}, \bibinfo{person}{Jonathan Ho}, \bibinfo{person}{Xi Chen}, \bibinfo{person}{Szymon Sidor}, {and} \bibinfo{person}{Ilya Sutskever}.} \bibinfo{year}{2017}\natexlab{}.
\newblock \bibinfo{title}{Evolution {Strategies} as a {Scalable} {Alternative} to {Reinforcement} {Learning}}.
\newblock
\newblock
\urldef\tempurl%
\url{https://doi.org/10.48550/arXiv.1703.03864}
\showDOI{\tempurl}
\newblock
\shownote{arXiv:1703.03864 [stat]}.


\bibitem[Schulman et~al\mbox{.}(2017a)]%
        {schulman_proximal_2017}
\bibfield{author}{\bibinfo{person}{John Schulman}, \bibinfo{person}{Filip Wolski}, \bibinfo{person}{Prafulla Dhariwal}, \bibinfo{person}{Alec Radford}, {and} \bibinfo{person}{Oleg Klimov}.} \bibinfo{year}{2017}\natexlab{a}.
\newblock \bibinfo{title}{Proximal {Policy} {Optimization} {Algorithms}}.
\newblock
\newblock
\urldef\tempurl%
\url{https://doi.org/10.48550/arXiv.1707.06347}
\showDOI{\tempurl}
\newblock
\shownote{arXiv:1707.06347 [cs]}.


\bibitem[Schulman et~al\mbox{.}(2017b)]%
        {schulman2017proximalpolicyoptimizationalgorithms}
\bibfield{author}{\bibinfo{person}{John Schulman}, \bibinfo{person}{Filip Wolski}, \bibinfo{person}{Prafulla Dhariwal}, \bibinfo{person}{Alec Radford}, {and} \bibinfo{person}{Oleg Klimov}.} \bibinfo{year}{2017}\natexlab{b}.
\newblock \bibinfo{title}{Proximal Policy Optimization Algorithms}.
\newblock
\newblock
\showeprint[arxiv]{1707.06347}~[cs.LG]
\urldef\tempurl%
\url{https://arxiv.org/abs/1707.06347}
\showURL{%
\tempurl}


\bibitem[Shuvaev et~al\mbox{.}(2024)]%
        {shuvaev_encoding_2024}
\bibfield{author}{\bibinfo{person}{Sergey Shuvaev}, \bibinfo{person}{Divyansha Lachi}, \bibinfo{person}{Alexei Koulakov}, {and} \bibinfo{person}{Anthony Zador}.} \bibinfo{year}{2024}\natexlab{}.
\newblock \showarticletitle{Encoding innate ability through a genomic bottleneck}.
\newblock \bibinfo{journal}{\emph{Proceedings of the National Academy of Sciences}} \bibinfo{volume}{121}, \bibinfo{number}{38} (\bibinfo{date}{Sept.} \bibinfo{year}{2024}), \bibinfo{pages}{e2409160121}.
\newblock
\urldef\tempurl%
\url{https://doi.org/10.1073/pnas.2409160121}
\showDOI{\tempurl}
\newblock
\shownote{Publisher: Proceedings of the National Academy of Sciences}.


\bibitem[Stanley et~al\mbox{.}(2019)]%
        {stanley_designing_2019}
\bibfield{author}{\bibinfo{person}{Kenneth~O. Stanley}, \bibinfo{person}{Jeff Clune}, \bibinfo{person}{Joel Lehman}, {and} \bibinfo{person}{Risto Miikkulainen}.} \bibinfo{year}{2019}\natexlab{}.
\newblock \showarticletitle{Designing neural networks through neuroevolution}.
\newblock \bibinfo{journal}{\emph{Nature Machine Intelligence}} \bibinfo{volume}{1}, \bibinfo{number}{1} (\bibinfo{date}{Jan.} \bibinfo{year}{2019}), \bibinfo{pages}{24--35}.
\newblock
\showISSN{2522-5839}
\urldef\tempurl%
\url{https://doi.org/10.1038/s42256-018-0006-z}
\showDOI{\tempurl}
\newblock
\shownote{Publisher: Nature Publishing Group}.


\bibitem[Stanley et~al\mbox{.}(2009)]%
        {stanley_hypercube-based_2009}
\bibfield{author}{\bibinfo{person}{Kenneth~O. Stanley}, \bibinfo{person}{David~B. D'Ambrosio}, {and} \bibinfo{person}{Jason Gauci}.} \bibinfo{year}{2009}\natexlab{}.
\newblock \showarticletitle{A {Hypercube}-{Based} {Encoding} for {Evolving} {Large}-{Scale} {Neural} {Networks}}.
\newblock \bibinfo{journal}{\emph{Artificial Life}} \bibinfo{volume}{15}, \bibinfo{number}{2} (\bibinfo{date}{April} \bibinfo{year}{2009}), \bibinfo{pages}{185--212}.
\newblock
\showISSN{1064-5462}
\urldef\tempurl%
\url{https://doi.org/10.1162/artl.2009.15.2.15202}
\showDOI{\tempurl}
\newblock
\shownote{Conference Name: Artificial Life}.


\bibitem[Stanley and Miikkulainen(2002)]%
        {stanley_evolving_2002}
\bibfield{author}{\bibinfo{person}{Kenneth~O. Stanley} {and} \bibinfo{person}{Risto Miikkulainen}.} \bibinfo{year}{2002}\natexlab{}.
\newblock \showarticletitle{Evolving {Neural} {Networks} through {Augmenting} {Topologies}}.
\newblock \bibinfo{journal}{\emph{Evolutionary Computation}} \bibinfo{volume}{10}, \bibinfo{number}{2} (\bibinfo{date}{June} \bibinfo{year}{2002}), \bibinfo{pages}{99--127}.
\newblock
\showISSN{1063-6560, 1530-9304}
\urldef\tempurl%
\url{https://doi.org/10.1162/106365602320169811}
\showDOI{\tempurl}


\bibitem[Stanley and Miikkulainen(2003)]%
        {stanley_taxonomy_2003}
\bibfield{author}{\bibinfo{person}{Kenneth~O. Stanley} {and} \bibinfo{person}{Risto Miikkulainen}.} \bibinfo{year}{2003}\natexlab{}.
\newblock \showarticletitle{A {Taxonomy} for {Artificial} {Embryogeny}}.
\newblock \bibinfo{journal}{\emph{Artificial Life}} \bibinfo{volume}{9}, \bibinfo{number}{2} (\bibinfo{date}{April} \bibinfo{year}{2003}), \bibinfo{pages}{93--130}.
\newblock
\showISSN{1064-5462, 1530-9185}
\urldef\tempurl%
\url{https://doi.org/10.1162/106454603322221487}
\showDOI{\tempurl}


\bibitem[Such et~al\mbox{.}(2018)]%
        {such_deep_2018}
\bibfield{author}{\bibinfo{person}{Felipe~Petroski Such}, \bibinfo{person}{Vashisht Madhavan}, \bibinfo{person}{Edoardo Conti}, \bibinfo{person}{Joel Lehman}, \bibinfo{person}{Kenneth~O. Stanley}, {and} \bibinfo{person}{Jeff Clune}.} \bibinfo{year}{2018}\natexlab{}.
\newblock \bibinfo{title}{Deep {Neuroevolution}: {Genetic} {Algorithms} {Are} a {Competitive} {Alternative} for {Training} {Deep} {Neural} {Networks} for {Reinforcement} {Learning}}.
\newblock
\newblock
\urldef\tempurl%
\url{https://doi.org/10.48550/arXiv.1712.06567}
\showDOI{\tempurl}
\newblock
\shownote{arXiv:1712.06567 [cs]}.


\bibitem[Taylor and Stone({[n.\,d.]})]%
        {taylor_transfer_nodate}
\bibfield{author}{\bibinfo{person}{Matthew~E Taylor} {and} \bibinfo{person}{Peter Stone}.} \bibinfo{year}{[n.\,d.]}\natexlab{}.
\newblock \showarticletitle{Transfer {Learning} for {Reinforcement} {Learning} {Domains}: {A} {Survey}}.
\newblock  (\bibinfo{year}{[n.\,d.]}).
\newblock


\bibitem[Thomson et~al\mbox{.}(2024)]%
        {thomson_understanding_2024}
\bibfield{author}{\bibinfo{person}{Sarah~L. Thomson}, \bibinfo{person}{Léni Le~Goff}, \bibinfo{person}{Emma Hart}, {and} \bibinfo{person}{Edgar Buchanan}.} \bibinfo{year}{2024}\natexlab{}.
\newblock \showarticletitle{Understanding {Fitness} {Landscapes} in {Morpho}-{Evolution} via {Local} {Optima} {Networks}}. In \bibinfo{booktitle}{\emph{Proceedings of the {Genetic} and {Evolutionary} {Computation} {Conference}}}. \bibinfo{publisher}{ACM}, \bibinfo{address}{Melbourne VIC Australia}, \bibinfo{pages}{114--123}.
\newblock
\showISBNx{9798400704949}
\urldef\tempurl%
\url{https://doi.org/10.1145/3638529.3654059}
\showDOI{\tempurl}


\bibitem[Vaswani et~al\mbox{.}(2023)]%
        {vaswani_attention_2023}
\bibfield{author}{\bibinfo{person}{Ashish Vaswani}, \bibinfo{person}{Noam Shazeer}, \bibinfo{person}{Niki Parmar}, \bibinfo{person}{Jakob Uszkoreit}, \bibinfo{person}{Llion Jones}, \bibinfo{person}{Aidan~N. Gomez}, \bibinfo{person}{Lukasz Kaiser}, {and} \bibinfo{person}{Illia Polosukhin}.} \bibinfo{year}{2023}\natexlab{}.
\newblock \bibinfo{title}{Attention {Is} {All} {You} {Need}}.
\newblock
\newblock
\urldef\tempurl%
\url{https://doi.org/10.48550/arXiv.1706.03762}
\showDOI{\tempurl}
\newblock
\shownote{arXiv:1706.03762 [cs]}.


\bibitem[Voudouris et~al\mbox{.}(2022)]%
        {voudouris_direct_2022}
\bibfield{author}{\bibinfo{person}{Konstantinos Voudouris}, \bibinfo{person}{Matthew Crosby}, \bibinfo{person}{Benjamin Beyret}, \bibinfo{person}{José Hernández-Orallo}, \bibinfo{person}{Murray Shanahan}, \bibinfo{person}{Marta Halina}, {and} \bibinfo{person}{Lucy~G. Cheke}.} \bibinfo{year}{2022}\natexlab{}.
\newblock \showarticletitle{Direct {Human}-{AI} {Comparison} in the {Animal}-{AI} {Environment}}.
\newblock \bibinfo{journal}{\emph{Frontiers in Psychology}}  \bibinfo{volume}{13} (\bibinfo{date}{May} \bibinfo{year}{2022}).
\newblock
\showISSN{1664-1078}
\urldef\tempurl%
\url{https://doi.org/10.3389/fpsyg.2022.711821}
\showDOI{\tempurl}
\newblock
\shownote{Publisher: Frontiers}.


\bibitem[Wang et~al\mbox{.}(2024)]%
        {wang_tensorized_2024}
\bibfield{author}{\bibinfo{person}{Lishuang Wang}, \bibinfo{person}{Mengfei Zhao}, \bibinfo{person}{Enyu Liu}, \bibinfo{person}{Kebin Sun}, {and} \bibinfo{person}{Ran Cheng}.} \bibinfo{year}{2024}\natexlab{}.
\newblock \showarticletitle{Tensorized {NeuroEvolution} of {Augmenting} {Topologies} for {GPU} {Acceleration}}. In \bibinfo{booktitle}{\emph{Proceedings of the {Genetic} and {Evolutionary} {Computation} {Conference}}} \emph{(\bibinfo{series}{{GECCO} '24})}. \bibinfo{publisher}{Association for Computing Machinery}, \bibinfo{address}{New York, NY, USA}, \bibinfo{pages}{1156--1164}.
\newblock
\showISBNx{9798400704949}
\urldef\tempurl%
\url{https://doi.org/10.1145/3638529.3654210}
\showDOI{\tempurl}


\end{thebibliography}

\clearpage
\appendix

\section{Implementation details}\label{app:hyperparams}
This appendix provides implementation details for the different methods evaluated in our study.

\subsection{NEAT}
We employed the tensorneat library~\citep{wang_tensorized_2024} with the hyperparameterers presented in Table~\ref{tab:neat_hyper}.
These values were tuned in the n-parity task without curriculum  and the same hyper-parameters were employed in all other tasks, with the exception of the population size and number of species that were reduced to 1024 and 20 in ecorobot environments to reduce computational complexity.

\begin{table}[H]
    \centering
    \begin{tabular}{|c|c|}
             \hline 
         Name & value  \\
         \hline
         population size & 5000 \\
         \hline 
           number of species & 50 \\
         \hline 
           add connection probability & 0.2 \\
         \hline 
        delete connection probability & 0.2 \\
         \hline 
          mutation connection & 0.1 \\      
                 \hline 
 mutation node & 0.1 \\
\hline 
          node deletion probability & 0.1 \\
 \hline 
architecture & feedforward \\
\hline
add node probability & 0.2 \\
                 \hline 
    \end{tabular}
    \caption{Hyperparameters for NEAT}
    \label{tab:neat_hyper}
\end{table}

\subsection{HyperNEAT}
We employed the tensorneat library~\citep{wang_tensorized_2024}.
For HyperNEAT we employ the same hyperparameters with NEAT for evolving the CPPN and the additional hyperparameters presented in Table~\ref{tab:hyperneat_hyper}.
HyperNEAT requires defining a neural network architecture with neurons having spatial locations on a two-dimensional plane (as this information is used by the CPPN).
We have employed networks with a single hidden layer (we did a hyperparameter search for different number of layers and neurons per layer in the n-parity task and did not observe any performance benefits but it is likely that further tuning the architecture will offer some performance gains in the ecorobot tasks.

\begin{table}[H]
    \centering
    \begin{tabular}{|c|c|}
             \hline 
         Name & value  \\
         \hline
         population size & 5000 \\
         \hline 
           number of species & 50 \\
         \hline 
           add connection probability & 0.2 \\
         \hline 
        delete connection probability & 0.2 \\
         \hline 
          mutation connection & 0.1 \\      
                 \hline 
 mutation node & 0.1 \\
\hline 
          node deletion probability & 0.1 \\
 \hline 
architecture & feedforward \\
\hline
add node probability & 0.2 \\
                 \hline 
    \end{tabular}
    \caption{Hyperparameters for HyperNEAT}
    \label{tab:hyperneat_hyper}
\end{table}

\begin{figure}
    \centering
    \includegraphics[width=0.9\linewidth]{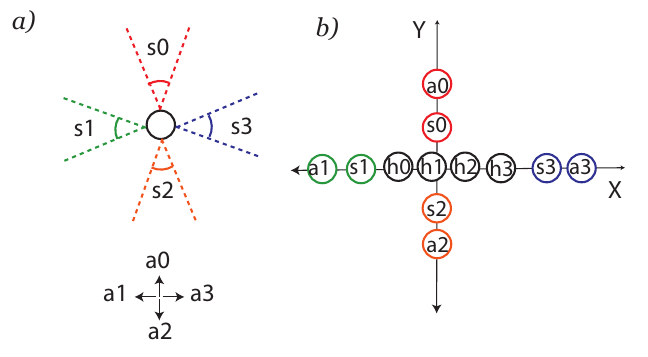}
    \caption{Illustration of how the substrate for HyperNEAT can be defined in ecorobot {\normalfont (a) SimpleROb can move in four direction using actions ($a0,a1,a2,a3$) and is equipped with four pie-slice sensors that point to the four corresponding directions ($s0,s1,s2,s3$). (b) We can place the neurons of the policy network on the substrate in the way that respects the geometry of the task. The location of hidden neurons instead is arbitrary (we indicate four hidden neurons for illustration purposes but employed 32 in our experiments}}
    \label{fig:hyperneat_substrate}
\end{figure}

\subsection{CMA-ES}
We employed the default hyper-parameters of evosax \cite{evosax2022github} for a feedforward neural network with 24 hidden neurons and all skip connections present.

\subsection{PPO} \label{app:PPO_details}
We employed the Brax implementation of PPO~\citep{freeman_brax_2021}.
We observed large benefits from tuning independently in the two benchmarks and expect further benefits from tuning separately on each task.
The largest benefit came from tuning the neural network architecture.
We present the hyperparameters for the two benchmarks in Tables~\ref{tab:ppo_hyper_stepping} and~\ref{tab:ppo_hyper_ecorobot}

\begin{table}[H]
    \centering
    \begin{tabular}{|c|c|}
             \hline 
         Name & value  \\
         \hline
                    architecture &  6 layers of 4 neurons \\

         number of updates per batch & 4 \\
         \hline 
           discounting & 1 \\
         \hline 
           learning rate &3e-4 \\
         \hline 
         number of minibatches & 32 \\
         \hline 
          batch size & 20498 \\      
                 \hline 
 entropy cost & 1e-2 \\
\hline 
           reward scaling & 1 \\

 \hline 
 unroll length & 1 \\
 \hline 
    \end{tabular}
    \caption{Hyperparameters for PPO and goal-conditioned PPO in stepping gates}
    \label{tab:ppo_hyper_stepping}
\end{table}

\begin{table}[H]
    \centering
    \begin{tabular}{|c|c|}
             \hline 
         Name & value  \\
         \hline
         number of updates per batch & 4 \\
         \hline 
                             architecture &  4 layers of 32 neurons \\
\hline 
           discounting & 0.97 \\
         \hline 
           learning rate &3e-4 \\
         \hline 
         number of minibatches & 32 \\
         \hline 
          batch size & 20498 \\      
                 \hline 
 entropy cost & 1e-2 \\
\hline 
           reward scaling & 10 \\

 \hline 
 unroll length & 5 \\
 \hline 
    \end{tabular}
    \caption{Hyperperameters for PPO and goal-conditioned PPO in ecorobot}
    \label{tab:ppo_hyper_ecorobot}
\end{table}

\subsection{MAP-Elites}
The algorithm used was an out-of-the-box implementation \citep{lim_accelerated_2022}, with the only change being the grid of the map set to consist of 50-by-50 tiles. This was done to best reflect the behavior descriptors which were the final x- and y-coordinates that the agent visited in the maze.
We report all relevant hyperparameters in Appendix \ref{tab:mapelites}.
\begin{table}[H]
    \centering
    \begin{tabular}{|c|c|}
                 \hline 
         Name & value  \\
         \hline 
                             architecture &  4 layers of 32 neurons \\

             \hline 
                             number of loops &  100 \\
            \hline 
                             number of iterations &  1000 \\

            \hline 
                             batch size &  100 \\
                        \hline 
                             centroids &  50-by-50 Euclidean grid \\

    \hline 
    \end{tabular}
    \caption{Hyperparameters for MAP-elites  }
    \label{tab:mapelites}
\end{table}

\end{document}